\documentclass[10pt,twocolumn,letterpaper]{article}

\usepackage[pagenumbers]{cvpr} 

\usepackage[table,xcdraw]{xcolor}
\usepackage[percent]{overpic}
\usepackage{xcolor}
\usepackage{tikz}

\usepackage{multirow}
\usepackage{array}

\newcolumntype{A}{>{\columncolor{blue!6}}c}
\newcolumntype{B}{>{\columncolor{green!6}}c}
\newcolumntype{C}{>{\columncolor{orange!8}}c}
\newcolumntype{D}{>{\columncolor{purple!6}}c}

\newcolumntype{E}{>{\columncolor{blue!6}}c}
\newcolumntype{F}{>{\columncolor{orange!8}}c}

\definecolor{cvprblue}{rgb}{0.21,0.49,0.74}
\usepackage[pagebackref,breaklinks,colorlinks,allcolors=cvprblue]{hyperref}
\usepackage{float}
\def\paperID{} 
\def\confName{CVPR}
\def\confYear{2026}

\title{Learning to Drive is a Free Gift: Large-Scale Label-Free Autonomy Pretraining from Unposed In-The-Wild Videos}



\author{%
Matthew Strong\textsuperscript{1,2\textdagger} \quad
Wei-Jer Chang\textsuperscript{1,3\textdagger} \quad
Quentin Herau\textsuperscript{1} \quad
Jiezhi Yang\textsuperscript{1}\\
Yihan Hu\textsuperscript{1} \quad
Chensheng Peng\textsuperscript{1,3\textdagger} \quad
Wei Zhan\textsuperscript{1,3\ddag}\\[0.4em]
{\textsuperscript{1}\,Applied Intuition \quad \textsuperscript{2}\,Stanford University \quad \textsuperscript{3}\,UC Berkeley}%
}

\begin{document}
\twocolumn[{%
\renewcommand\twocolumn[1][]{#1}%
\maketitle
\vspace{-0.5cm}
\begin{center}
    \centering
    \captionsetup{type=figure}
    \includegraphics[width=\textwidth]{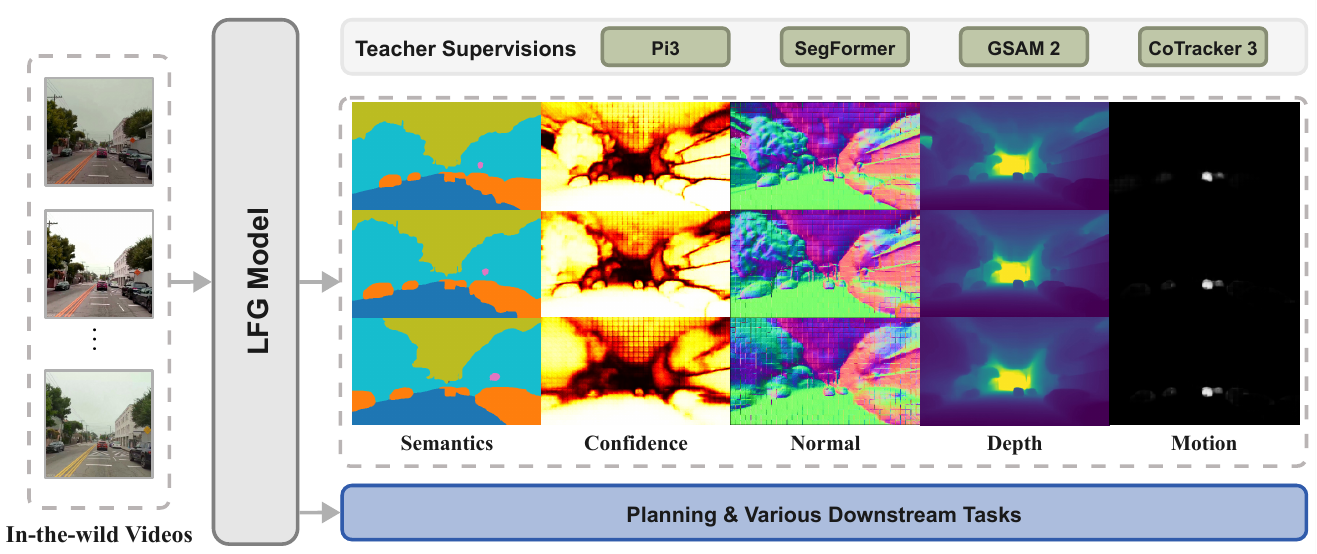}
\captionof{figure}{LFG learns a unified pseudo-4D representation of geometry, semantics, motion, and short-term future evolution directly from unposed, unlabeled single-view driving videos. A single feedforward encoder processes observed frames and produces temporally consistent predictions of 3D point maps, camera poses, semantic layouts, confidence, and motion masks for both current and future frames.
}
    \label{fig:teaser}
\end{center}%
}]

\renewcommand{\thefootnote}{}
\footnotetext{\hspace{-1.8em}\textsuperscript{\textdagger}\,Work done as intern at Applied Intuition.\\
\textsuperscript{\ddag}\,Correspondence: \texttt{wei.zhan@applied.co}}
\renewcommand{\thefootnote}{\arabic{footnote}}

\begin{abstract}

Ego-centric driving videos available online provide an abundant source of visual data for autonomous driving, yet their lack of annotations makes it difficult to learn representations that capture both semantic structure and 3D geometry. Recent advances in large feedforward spatial models demonstrate that point maps and ego-motion can be inferred in a single forward pass, suggesting a promising direction for scalable driving perception. We therefore propose a label-free, teacher-guided framework for learning autonomous driving representations directly from unposed videos. Unlike prior self-supervised approaches that focus primarily on frame-to-frame consistency, we posit that safe and reactive driving depends critically on temporal context. To this end, we leverage a feedforward architecture equipped with a lightweight autoregressive module, trained using multi-modal supervisory signals that guide the model to jointly predict current and future point maps, camera poses, semantic segmentation, and motion masks. Multi-modal teachers provide sequence-level pseudo-supervision, enabling LFG to learn a unified pseudo-4D representation from raw YouTube videos without poses, labels, or LiDAR. The resulting encoder not only transfers effectively to downstream autonomous driving planning on the NAVSIM benchmark, surpassing multi-camera and LiDAR baselines with only a single monocular camera, but also yields strong performance when evaluated on a range of semantic, geometric, and qualitative motion prediction tasks. These geometry and motion-aware features position LFG as a compelling video-centric foundation model for autonomous driving. Check out our project page at \url{https://lfg-ai.github.io/}.
\end{abstract}    
\section{Introduction}
\label{sec:intro}

In-the-wild, ego-centric driving videos available online provide an abundant source of visual data for driving, yet their lack of annotations makes it difficult to learn representations that capture both semantic, temporal structure and 3D geometry. Inspired by the recent success of GPT-style models~\cite{openai2023gpt4, deepmind2023gemini} and DINOv3~\cite{simeoni2025dinov3} trained on massive unlabeled internet corpora, a natural question arises: can we similarly leverage large amounts of raw video to learn geometry and motion aware features for autonomy?

 Recent research in autonomy has shown that scaling up improves performance~\cite{hu2023planning,li2024hydramdp,dauner2024navsim}, yet most approaches still rely heavily on \emph{labeled} data in the form of expert actions, LiDAR scans, odometry, and semantic annotations. Meanwhile, in-the-wild driving videos are abundant and capture a wide range of visual conditions and traffic situations. Although these videos provide only RGB information, they contain rich visual and motion cues that can be learned. If we aim to build scalable autonomy models capable of producing expressive and actionable representations, they should benefit from large-scale pretraining on unlabeled images and videos.
 
This motivates the goal of learning structure and motion directly from video. Feedforward 3D reconstruction models already demonstrate that it is possible to estimate camera poses and point maps from unposed image sequences using a single forward pass~\cite{wang2025vggt,wang2025pi3}. Egocentric driving videos provide ideal data for such models, as consecutive frames naturally encode geometry and ego-motion, even with sparse viewpoints. Yet for autonomous driving, a model must ultimately do more: beyond reconstructing the present, \textit{it must predict future motion and geometry}.

Motivated by findings that humans make low-level driving decisions from only a short motion history, we extend the feedforward reconstruction model $\pi^3$~\cite{wang2025pi3} to predict future geometry, confidence, and motion. Our model is trained using signals from multiple large-scale models trained on unposed data, which provide complementary cues for geometry, motion, and semantics. By integrating these cues and incorporating segmentation and motion components, the student model learns from in-the-wild driving videos to produce a pseudo-4D output that captures scene structure together with the motion of dynamic agents.

We introduce LFG -- \textbf{L}earning to drive is a \textbf{F}ree \textbf{G}ift -- a label-free, teacher-guided approach for learning such representations from vision alone. We formulate future prediction as a next-token prediction problem over geometry, motion, and semantic features. A lightweight autoregressive transformer is added after the reconstruction aggregator, enabling a student model trained on a subset of views to benefit from stronger models with access to the full sequence. Supervision comes from several specialized teachers—SegFormer~\cite{xie2021segformer} for semantics, SAM2~\cite{li2024sam2} and CoTracker3~\cite{karaev2024cotracker3} for motion cues —each used in a way that best leverages its strengths on unlabeled driving video.

Unlike large world models that still require a degree of supervised labels~\cite{hafner2023mastering,ho2022video,blattmann2023stable,kulhanek2019world}, LFG focuses on a short-horizon, feedforward formulation that sets a new standard among geometry-aware models for autonomy. On the NAVSIM planning benchmark~\cite{dauner2024navsim}, LFG achieves state-of-the-art performance using \emph{only a single front-camera view}, outperforming multi-view and BEV-based methods such as UniAD~\cite{li2023uniad} and HydraMDP~\cite{li2024hydramdp}, which rely on multiple cameras, LiDAR, or both. LFG pretraining also provides strong sample efficiency: with only 10\% labeled data, it achieves competitive planning performance, underscoring the value of large-scale training on unlabeled driving video. Beyond planning, LFG produces geometry- and motion-aware features that transfer effectively to tasks spanning semantics, 3D structure, and decision making, underscoring its broader applicability as a backbone for next-generation autonomous driving systems.

\paragraph{Our main contributions are as follows:}
\begin{itemize}
    \item We propose \textbf{LFG}, a label-free, video-centric pre-training framework that learns geometry-, motion-, and semantics-aware representations directly from unposed, single-view driving videos.
    \item We design a unified architecture built on a pretrained encoder and a causal autoregressive module, enabling short-horizon prediction of point maps, camera poses, semantic layouts, confidence maps, and motion masks under multiple teacher-guided supervision.
    \item We demonstrate that LFG serves as a strong foundation for autonomy: it achieves state-of-the-art planning performance using only a single front camera, exhibits compelling data efficiency, and transfers effectively across semantic, geometric, and motion tasks. We emphasize that the novelty of LFG lies more within the pretraining paradigm than the model itself.
\end{itemize}

\section{Related Work}
\label{sec:related_work}


\begin{figure*}
    \centering
    \includegraphics[width=1\linewidth]{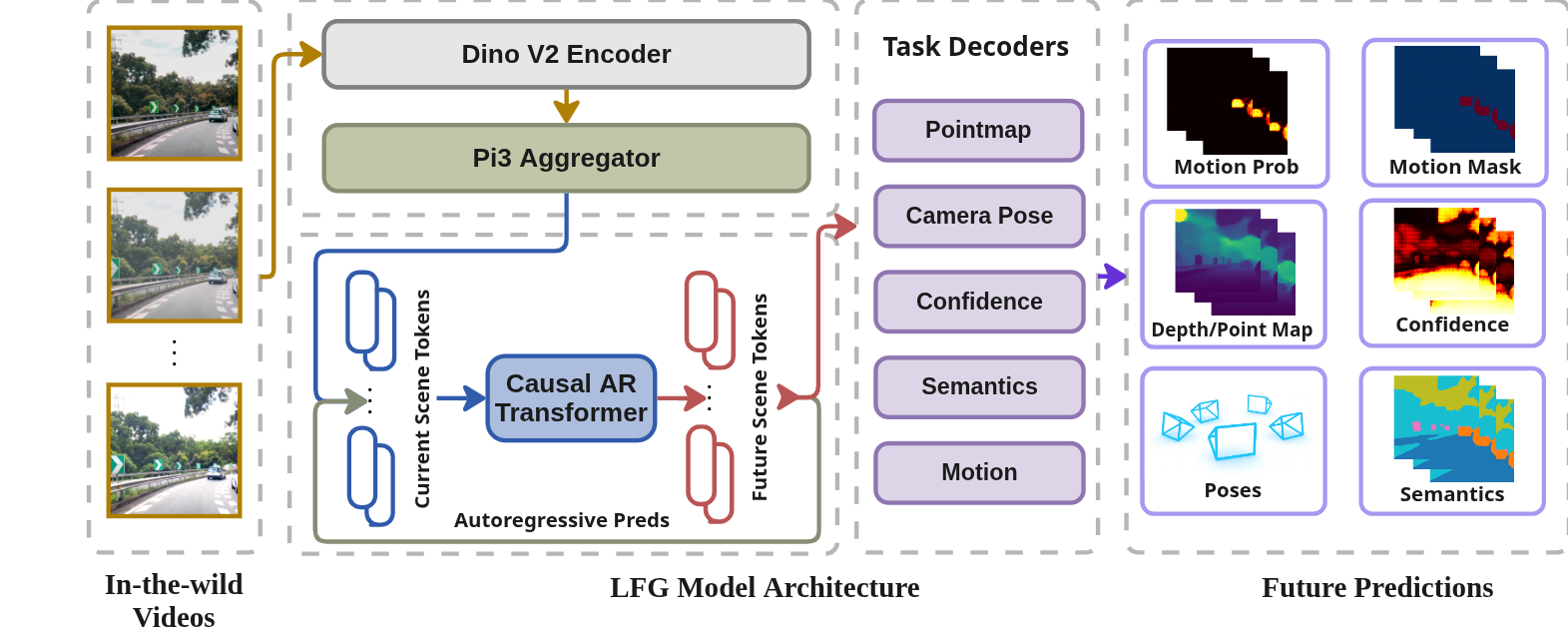}
    \caption{
\textbf{LFG architecture.}
Starting from unposed single-view driving clips, a pretrained $\pi^3$ backbone encodes $N$ observed frames into latent scene tokens.
A lightweight causal autoregressive transformer rolls out $M$ future tokens, which a shared decoder maps to point maps, camera poses, semantic segmentation, confidence maps, and motion masks for all $N{+}M$ frames.
Multi-modal teachers provide pseudo-supervision, enabling LFG to learn a unified pseudo-4D representation that transfers effectively to downstream planning.
}
    \label{fig:lfg_arch}
\end{figure*}



\textbf{Pretraining for Autonomous Driving.}
Pretraining for autonomous driving has only recently gained traction. Early self-supervised pretraining work such as SelfD~\cite{zhang2022selfd} and ACO~\cite{zhang2022learning} demonstrated that large-scale in-the-wild driving videos can provide supervisory signals for learning semantic and geometric priors without human labels. PPGeo~\cite{wu2023policy} further explored geometry-oriented pretraining using photometric and consistency-based objectives to learn depth and ego-motion. ViDAR~\cite{yang2024visual} proposes to use future point-cloud prediction from historical camera inputs as a unified pretext task. UniPAD~\cite{yang2024unipaduniversalpretrainingparadigm} introduces a self-supervised learning paradigm that uses 3D volumetric differentiable rendering to implicitly encode continuous 3D structures. VisionPAD~\cite{zhang2025visionpadvisioncentricpretrainingparadigm} focuses on vision-centric algorithms by leveraging efficient 3D Gaussian Splatting and a multi-frame photometric consistency objective to reconstruct multi-view representations using only images. However, these approaches largely rely on frame-to-frame consistency losses that implicitly assume static scenes, limiting their ability to capture dynamic objects that are central to real driving environments. In contrast, our method is pretrained directly on unlabeled driving video by explicitly modeling \emph{dynamic} geometry, motion cues, and scene semantics, yielding a dense 4D representation that better captures the structure and dynamics of real-world driving.





\textbf{Geometry-aware vision backbones for driving.}
Classical 3D reconstruction pipelines in autonomy rely on Structure-from-Motion (SfM) and Multi-View Stereo (MVS)~\cite{schonberger2016structure}, often combined with LiDAR, to triangulate scene points and build dense maps for localization. While effective, these methods are typically tailored per scene and are not naturally suited as general-purpose backbones for large-scale video pretraining. In contrast, recent feedforward approaches~\cite{wang2025vggt,wang2025pi3,murai2025mast3r,tian2025drivingforward,lirig3r} amortize reconstruction by predicting point maps, confidence maps, and camera poses for unposed image sequences in a single pass, making them attractive as scalable, geometry-aware backbones for driving. LFG belongs to this family but focuses on \emph{temporal} understanding, producing a pseudo-4D representation of dynamic driving scenes that is well suited for downstream planning and perception.

\section{Method}
\label{sec:method}

We introduce LFG (shown in \cref{fig:lfg_arch}), a method for learning a powerful driving-vision model from unposed and unlabelled single view Youtube videos. 


\begin{figure}
    \centering
    \includegraphics[width=1\linewidth]{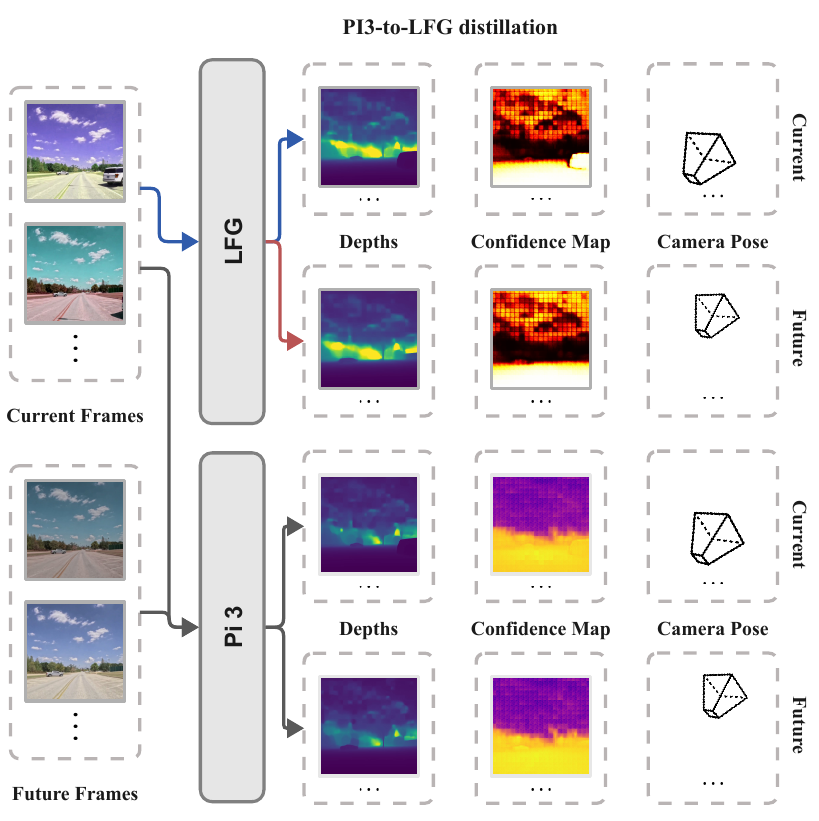}
    \caption{
\textbf{$\pi^3$-to-LFG distillation.}
We transfer geometric knowledge from the pretrained $\pi^3$ teacher to LFG by supervising point maps, confidence maps, and camera poses for all observed and future frames. 
While the teacher has access to the full sequence, the student sees only the first $N                                                                                                                                                                                                                                                                                                                                                                                                                                                                                                                                                                                                                                                                                                                                                                                                                                                                                                                                                                                                                                                                                                                                                                                                                                                                                                                                                                                                                                                                                                                                                                                                                                                                                                                                                                                                                                                                                                                                                                                                                                                                                                                                                           $ frames and must predict both current and future geometry, enabling LFG to learn temporally consistent scene structure and future ego-motion from partial observations.
}
    \label{fig:teacher}
\end{figure}

\subsection{Problem Formulation}
\label{sec:problem}

We consider the case of learning to drive, where a large parameterized model is given a consecutive sequence $(I_t)_{t=1}^N$ of $N$ ego-centric RGB images $I_t \in \mathbb{R}^{3 \times H \times W}$, in a variety of driving scenes. The goal is to efficiently predict scene information that is useful for autonomous driving. We posit that this includes both current and short-horizon future information. We posit that this includes current \textit{and} future information in the recent future. Such a model should predict $\mathcal{O}$ relevant modalities of scene information, as well as the future $M$ frames of the scene. Inspired by prior work in label-free pretraining and world models for driving, we choose to predict the following outputs.

LFG processes in-the-wild video through a pretrained encoder and a causal autoregressive transformer to jointly predict current and short-horizon future scene geometry, semantics, and motion.
First, our model should predict \textbf{point maps} for the ego-view camera over time. $(P_t)_{t=1}^{N+M}, \quad P_t : \mathcal{I}(I_t) \rightarrow \mathbb{R}^3$, where $\mathcal{I}(I_t)$ maps each pixel in $I_t$ to $P_t(\mathbf{y}) \in \mathbb{R}^3$, which is the 3D world point corresponding to pixel $\mathbf{y}$ at time $t$.

Second, our model predicts \textbf{camera poses}: \((T_t)_{t=1}^{N+M}, \; T_t \in \mathbb{R}^{4 \times 4}\), where each \(p_t\) is a full \(4 \times 4\) homogeneous transformation matrix encoding both rotation and translation. 
Such poses define the ego-motion trajectory of the camera and enable mapping all predicted local 3D points into a shared world coordinate frame.

Third, our model predicts \textbf{semantic segmentation} with 7 classes: $(S_t)_{t=1}^{N+M}, \quad S_t \in \mathbb{R}^{7 \times H \times W}$,
where each pixel’s one-hot vector \(S_t(\mathbf{y}) \in \mathbb{R}^{7}\) encodes the semantic category (e.g., road, vehicle, pedestrian, building, vegetation, sky, and background). These semantic predictions provide a semantic, structured understanding of the scene, which we consider to be useful for the downstream task.

We also predict 
\textbf{confidence maps}
$(C_t)_{t=1}^{N+M}, \quad C_t : \mathcal{I}(I_t) \rightarrow [0,1],$
which quantifies the reliability of each pixel’s 3D prediction.

Finally, our model should predict \textbf{motion masks}
$(M_t)_{t=1}^{N+M}, \quad M_t : \mathcal{I}(I_t) \rightarrow [0,1],$
indicating which regions in the image correspond to independently moving objects (e.g., other vehicles, pedestrians) as opposed to the static environment. 
The motion masks help disentangle dynamic from static components of the scene, which can be used for downstream tasks, such as dynamic 4D Gaussian Splatting.

In total, the model predicts all outputs:
\[
    \mathcal{O} = \{ (P_t, T_t, S_t, C_t)_{t=1}^{N+M}, \; (M_t)_{t=1}^{N+M} \},
\]

All modalities are learned jointly in an end-to-end fashion from video, with the assistance of robust teachers, promoting shared representations of geometry, semantics, and motion relevant to autonomous driving.

\begin{figure}
    \centering
    \includegraphics[width=0.9\linewidth]{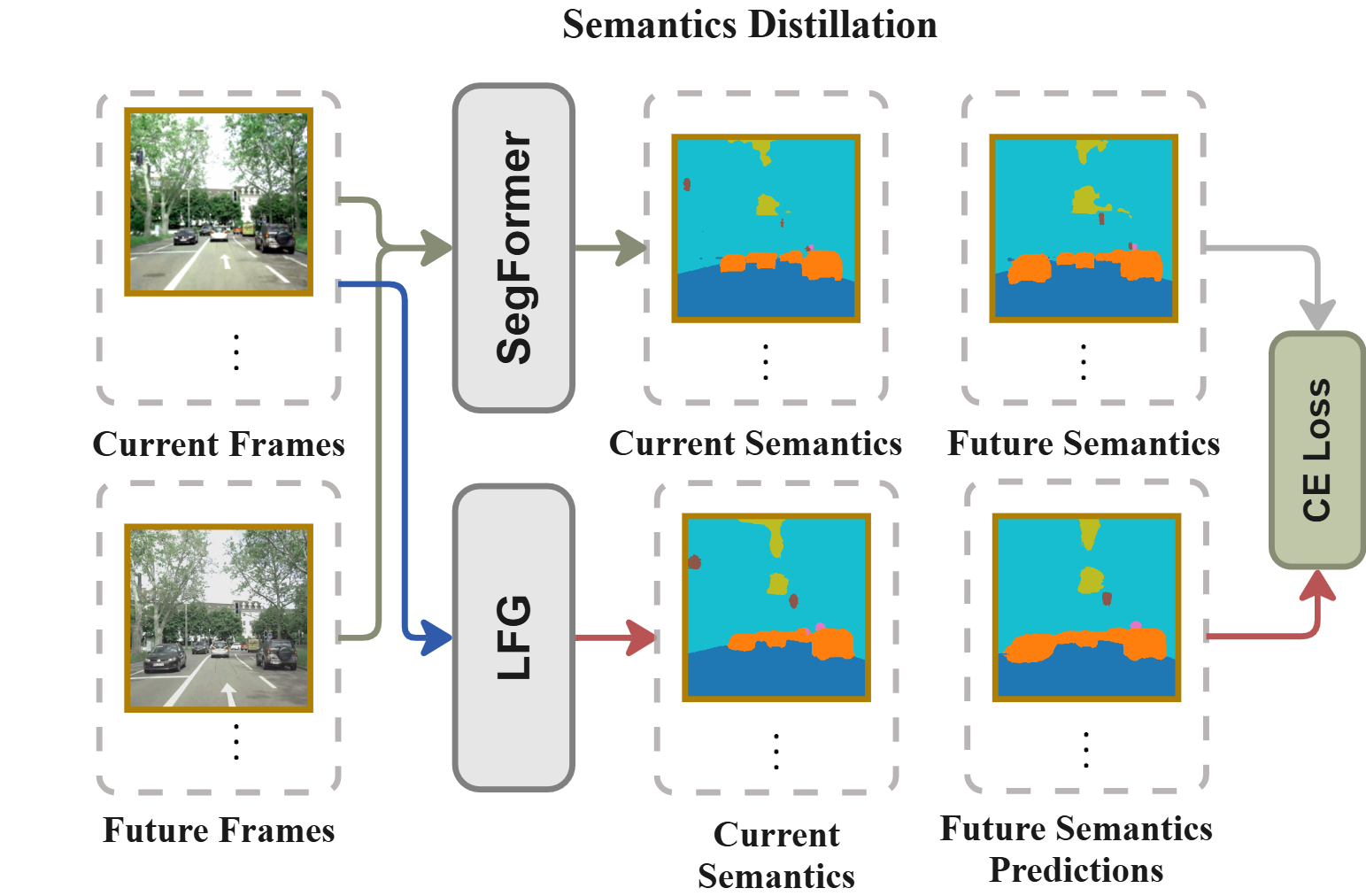}
    \caption{
\textbf{Semantic distillation.}
A pretrained SegFormer teacher, trained on Cityscapes, provides soft semantic pseudo-labels for each frame. 
LFG predicts semantic maps for both observed and future frames using only the first $M$ inputs, learning temporally consistent scene semantics through teacher–student supervision aligned with the model’s geometric features.
}
    \label{fig:semantic_head}
\end{figure}

\subsection{Architecture.}

Our model (\cref{fig:lfg_arch}) is built on top of the $\pi^3$,  \cite{wang2025pi} model, which is a purely feedforward model that predicts point maps, confidence maps, and camera poses from a series of unposed images. Contrary to prior work in VGGT \cite{wang2025vggt}, $\pi^3$ does not rely on a fixed referenced view, and is trained on more \textit{dynamic} datasets, making it a suitable starting point for LFG. To receive the benefits of the pretrained $\pi^3$, we propose some simple additions on top of the model.

First, we propose to add a \textbf{causal attention autoregressive transformer} after $\pi^3$'s alternating attention module or encoder. 
Let the output of the $\pi^3$ encoder be a sequence of latent scene tokens $\mathbf{Z}_{1:N}$, 
where $N$ is the number of observed frames.
The autoregressive transformer $\mathcal{T}_{\mathrm{AR}}$ takes these tokens as input and causally predicts additional latent tokens for $M$ future frames, producing $\mathbf{Z}_{1:N+M} = \mathcal{T}_{\mathrm{AR}}(\mathbf{Z}_{1:N})$.
Each newly generated token sequence $\mathbf{Z}_{N+1:N+M}$ represents latent scene features for unobserved frames, which are decoded into 3D point maps, confidence maps, camera poses, semantic maps, and motion masks. Our causal formulation ensures that each predicted future frame can attend to past and observed frames, but not to future frames, enforcing a forward-only information flow. The semantic and motion outputs are initialized from the point decoder and their respective heads, allowing the model to leverage shared geometric features while predicting scene semantics and dynamics.

\paragraph{$\pi^3$ Teacher.}
We employ a teacher \(\pi^3\) model, seen in Fig. \ref{fig:teacher}, that has access to \(N + M\) frames from the unlabeled OpenDV dataset~\cite{yang2024driveagi}. The teacher outputs supervision signals in the form of point maps, confidence maps, and camera poses for \textbf{all} \(N + M\) frames. 
Our student model only observes the first \(N\) frames and must predict both the observed (\(N\)) and future (\(M\)) outputs. Specifically, the student, LFG, predicts:
\[
\{\mathbf{P}_t, \mathbf{C}_t, \mathbf{T}_t\}_{t=1}^{N+M},
\]
where \(\mathbf{P}_t\) denotes the point map, \(\mathbf{C}_t\) the confidence map, and \(\mathbf{T}_t\) the camera pose at frame \(t\). While this method is not self-supervised as compared to other works, it forces LFG to predict future information, namely the future ego motion, confidence, and geometric updates of the scene.

\subsection{Semantic Head}
\label{sec:semantic_head}

To enable semantic understanding of the scene, our model includes a \textbf{semantic head} (Fig. \ref{fig:semantic_head}) that predicts dense per-pixel class probabilities for each camera and timestep. 
Given the input sequence of $N$ images $(I_t)_{t=1}^{N}$, the semantic head outputs a corresponding sequence of current and future semantic maps $(S_t)_{t=1}^{N+M}, \quad S_t \in [0,1]^{C_s \times H \times W}$. Since ground-truth semantic labels are unavailable for all frames, we turn to a simple \textit{teacher–student} training strategy. 
A pretrained \textbf{SegFormer} model $\Phi_{\text{seg}}$, trained on the Cityscapes dataset~\cite{Cordts2016Cityscapes}, serves as the teacher network. 
For each image $I_t$, we obtain pseudo-labels: $\hat{S}_t = \Phi_{\text{seg}}(I_t)$. These pseudo-labels act as soft supervision targets for the semantic head. The SegFormer teacher is given access to all frames, while LFG has to predict the current and future segmentation predictions.

\begin{figure}
    \centering
    \includegraphics[width=1\linewidth]{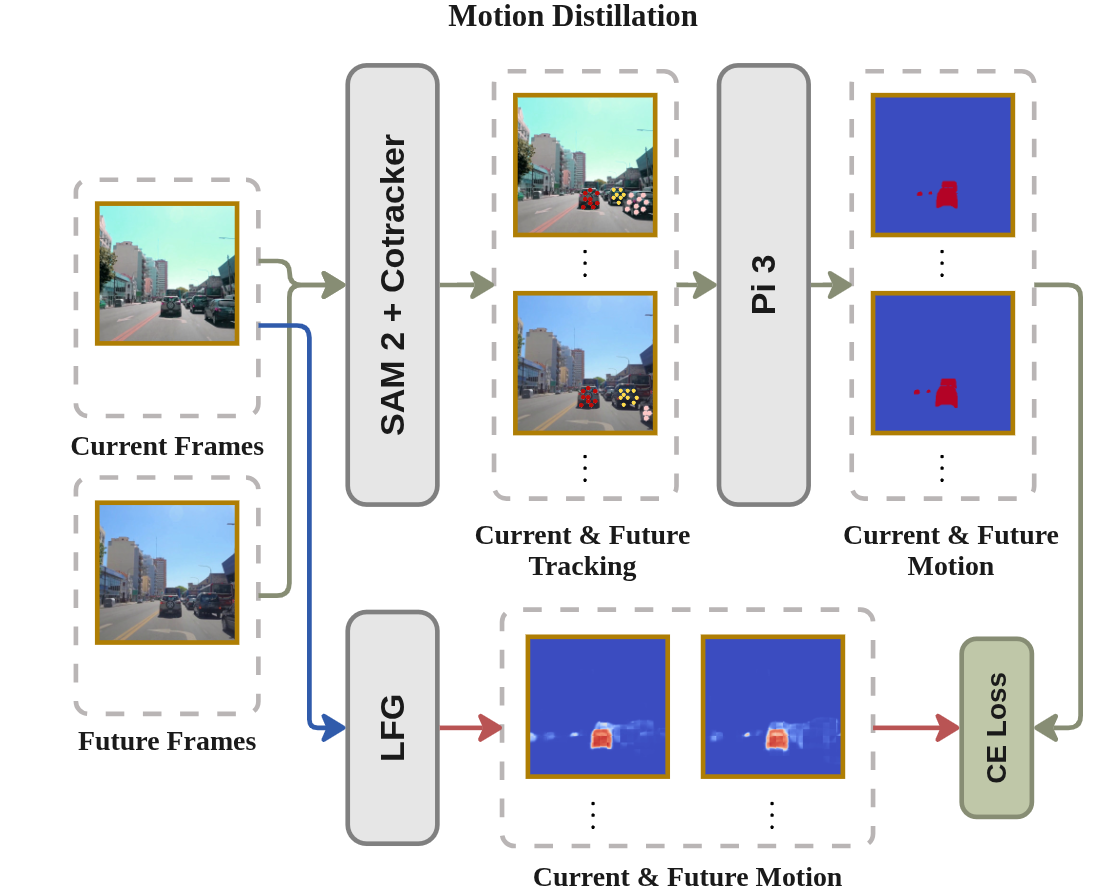}
    \caption{
\textbf{Motion mask generation pipeline.}
We first detect human and vehicle instances in the first frame using Grounded SAM2, then track their 2D trajectories across time with CoTracker3.
Using teacher $\pi^3$ point maps, tracked pixels are backprojected into 3D and per-instance 3D displacements are measured over the sequence.
Instances whose motion exceeds a threshold for at least $K_{\min}$ frames are labeled as dynamic, and their masks are rasterized into dense per-pixel motion masks $\mathbf{M}_t$, which supervise the motion head.
}
    \label{fig:motion_figure}
\end{figure}

\subsection{Motion Head}

Our motion head in Fig. \ref{fig:motion_figure} predicts per-pixel motion masks that identify dynamic regions within a scene. Since explicit motion annotations are unavailable, we generate pseudo ground-truth (pseudo-GT) labels in a fully feedforward, label-free manner.

We begin by segmenting human and vehicle instances from the first frame by using an off-the-shelf segmentation model, Grounded SAM2 \cite{ren2024grounded}, which produces a list of tracked mask instances per object. For each detected object, we track its 2D motion across frames using \textit{CoTracker3} \cite{karaev2025cotracker3}, which provides dense correspondences in image space: $\mathbf{u}_{t}^{(i)} = \mathrm{CoTracker3}(\mathbf{I}_1, \ldots, \mathbf{I}_T, i)$, where \(\mathbf{u}_{t}^{(i)}\) denotes the 2D tracked keypoints of object \(i\) at frame \(t\).

Next, we employ the teacher \(\pi^3\) model to obtain corresponding 3D point maps for each frame. For each object instance \(i\), we backproject the tracked 2D points into 3D using \(\mathbf{P}_t\), and measure the temporal displacement of the mean 3D position:
\[
d_t^{(i)} = \left\| \bar{\mathbf{p}}_{t+1}^{(i)} - \bar{\mathbf{p}}_{t}^{(i)} \right\|_2,
\]
where \(\bar{\mathbf{p}}_{t}^{(i)}\) is the mean 3D position of the object at time \(t\). An object is considered \textit{dynamic} if its displacement exceeds a motion threshold \(\tau_{\text{motion}}\) for at least \(K_{\min}\) frames.
Finally, we convert instance-level motion indicators \(m^{(i)}\) into dense motion masks \(\mathbf{M}_t \in [0,1]^{H \times W}\) that serve as supervision for the motion head.

\subsection{Losses}

Our training objective combines multiple task-specific loss terms that jointly supervise segmentation, geometry, motion, and camera pose estimation. The total training loss is:
\begin{equation}
\begin{aligned}
\mathcal{L}_{\text{total}} =\;
& \mathcal{L}_{\text{current}}
+ \lambda_{\text{future}}\,\mathcal{L}_{\text{future}}.
\end{aligned}
\end{equation}

\begin{equation}
\begin{aligned}
\mathcal{L}_{\text{current/future}} =\;
& \lambda_{\text{seg}}\,\mathcal{L}_{\text{seg}}
+ \lambda_{\text{pose}}\,\mathcal{L}_{\text{pose}} \\
& + \lambda_{\text{point}}\,\mathcal{L}_{\text{point}}
+ \lambda_{\text{motion}}\,\mathcal{L}_{\text{motion}}.
\end{aligned}
\end{equation}

\subsubsection{Segmentation Loss.}
We use a weighted BCE loss for semantic segmentation, where we use class-specific weight to handle class imbalance. Please see the supplementary material for additional details.

\subsubsection{Pose Loss.}
Following the $\pi^3$ formulation, we supervise the predicted camera poses using relative pose consistency across frame pairs. For any two frames $(i,j)$, we construct relative transformations $(\Delta \mathbf{R}_{i\leftarrow j}, \Delta \mathbf{t}_{i\leftarrow j})$ from the student predictions and compare them against teacher-provided targets $(\widehat{\Delta \mathbf{R}}_{i\leftarrow j}, \widehat{\Delta \mathbf{t}}_{i\leftarrow j})$. The overall loss combines rotation and translation terms:
\[
\mathcal{L}_{\text{pose}}
= \mathcal{L}_{\text{rot}} \;+\; \lambda_{\text{trans}}\, \mathcal{L}_{\text{trans}}.
\]
The rotation term penalizes geodesic distance on $\mathrm{SO}(3)$ between predicted and target relative rotations, while the translation term uses a robust regression loss (Huber) on relative translations to handle scale variation and outliers. This formulation enforces multi-frame pose consistency and stabilizes predictions over time.

\subsubsection{Confidence Loss.}
The confidence map estimates the reliability of each predicted 3D point. We supervise it using a binary target derived from the point-map reconstruction error: pixels whose point error falls below a threshold are treated as high-confidence, and others as low-confidence. We apply a binary cross-entropy loss to this target.

\subsubsection{Point Map Loss.} 
We supervise the predicted 3D point maps using a scaled $L_1$ loss to account for varying scene scales:
\[
\mathcal{L}_{\text{point}} = \alpha \, \|\mathbf{P} - \widehat{\mathbf{P}}\|_1,
\]
where $\mathbf{P}$ and $\widehat{\mathbf{P}}$ denote the predicted and target point maps, respectively, and $\alpha$ is a learned or fixed scaling factor that normalizes for scene scale. 
This formulation encourages accurate 3D reconstruction while remaining robust to the absolute magnitude of the scene, analogous to the Huber-based translation loss used for relative camera motion, where we also apply a scale.

\subsubsection{Motion Loss.}
The motion head is trained with a binary cross-entropy loss between the model prediction (LFG) and the pseudo ground-truth (GT):
\[
\mathcal{L}_{\text{motion}}
= - \sum \big[ M^{\text{GT}} \log M^{\text{LFG}} + (1 - M^{\text{GT}})\log (1 - M^{\text{LFG}}) \big].
\]

\subsubsection{Future Frame Weighting.}
To emphasize the model’s ability to predict beyond observed frames, we apply a temporal weighting factor \(\omega_t\) to all losses on future frames, keeping $\omega$ fixed:
\[
\mathcal{L}_{\text{future}} = 
\sum_{t=M+1}^{N+M} \omega \, \mathcal{L}_t,
\quad \text{with } \omega > 1.
\]
This encourages accurate extrapolation of geometry and motion into the future time steps.

Together, these terms ensure LFG spatially and semantically understands the scene, as well as how the scene will evolve in a recent future time window. By nature, LFG exhibits generative qualities in its autoregressor; however, we assert that this is needed for next frames prediction.

\subsection{Training}
We train LFG in three stages. The first stage ensures that LFG can predict \textit{future} geometry and pose autoregressively. This provides the autoregressive transformer a strong initialization to train the segmentation head, while not having to relearn future geometry and motion. Finally, we train on the motion masks, initialized from the point decoder. In each stage, LFG is trained end-to-end. We exclusively use the OpenDV Driving Youtube Daatset, and opt for a subset of it, consisting of approximately 2 million samples across varied driving conditions, scenes, traffic and external driver/pedestrian situations. We train our model on $2$, $5$, and $10$ Hz frames (without any conditioning LFG on frequency of input frames) to improve robustness.

\subsection{Fine-tuning for Planning}
\label{sec:finetuning}
With a strong pretrained encoder that captures temporal and spatial scene structure from sequential images, we now demonstrate how this representation benefits downstream planning. We fine-tune on the NAVSIM planning benchmark \cite{dauner2024navsim} using only front-view camera inputs over three consecutive frames to predict future trajectories in complex driving scenarios.

The pretrained image encoder backbone is kept frozen and, for each frame, outputs high-dimensional \textbf{autonomy tokens} that encode the ego vehicle's motion state and surrounding context. We run LFG to produce the \textit{future tokens} for learning. These per-frame features are aggregated and passed to a lightweight multi-modal \textbf{anchor-based trajectory decoder} that directly predicts multiple candidate trajectories in a single forward pass, similar to~\cite{liao2025diffusiondrive} but without any diffusion or iterative refinement. The decoder attends from autonomy features to trajectory anchors and across trajectory modes, then outputs confidence scores and coordinate offsets, selecting the highest-confidence mode as the final plan.

This simple yet effective fine-tuning strategy allows the planner to directly leverage the pretrained temporal representation for the planning task, leading to strong gains in data efficiency. In our experiments (Sec.~\ref{sec:exp_plan}), we show that this strong pretrained encoder substantially improves planning performance and data efficiency compared to state-of-the-art models that utilize multi-view or LiDAR inputs, as well as other pretrained encoders \cite{wu2023ppgeo}.




\section{Experiments}
\label{sec:experiments}

\subsection{Implementation and Training}

\subsubsection{Model and Pretraining.}  
We implement our model by closely following the architecture of the original $\pi^3$ backbone, which contains approximately 1 billion parameters. In total, LFG contains 1.45B parameters, and runs at 5Hz on an NVIDIA RTX 5090 GPU. The image encoder is initialized from a DINOv2-pretrained backbone, and we directly follow the $\pi^3$ alternating attention module. The point, confidence, and camera heads are frozen. The semantic and motion mask head are initialized from the point head. Our causal autoregressive transformer consists of 4 layers with 8 attention heads and a dropout rate of 0.1. It takes the latent scene tokens from the $\pi^3$ encoder and autoregressively predicts future frame tokens, which are then decoded to point maps, semantic maps, confidence maps, camera poses, and motion masks.

\subsubsection{Training Setup.}  
We train the model using the AdamW optimizer with a base learning rate of $10^{-4}$.  A linear warmup schedule is used for the first 500 steps, starting from $0.1 \times$ the base learning rate and increasing to the full learning rate. After warmup, we apply cosine annealing over the remaining training steps. Gradients are clipped to a maximum norm of 1.0, and mixed-precision training (BF16) is enabled. We perform gradient accumulation to increase batch size. We also randomly apply color jittering, Gaussian blur, and grayscale augmentation to the frames of the student LFG, while letting the teacher receive unaugmented images. We train on 32 A100  GPUs for 40,000 iterations. The model is trained using a combination of losses, including scaled $L_1$ for 3D points ($\alpha = 1.0$), Huber loss for camera translation ($0.1$), confidence loss ($0.05$), segmentation loss ($1.0$), and motion loss ($1.0$). To emphasize accurate prediction of future frames, we apply a weight of $ \omega = 10.0$ to the corresponding losses.  Finally, we normalize \textit{all geometric outputs} to ensure stable learning during training.

\subsection{Results}
We evaluate LFG on a suite of downstream tasks that jointly probe semantics, geometry, motion, and decision making. Concretely, we consider (i) semantic segmentation, (ii) depth, point map, and camera pose prediction, and (iii) encoder-only downstream benchmarks, planning. We additionally provide qualitative motion visualizations. These tasks allow us to assess both the quality of the learned scene representation and its usefulness as a backbone for autonomous driving.

\begin{figure}[h]
    \centering
    \includegraphics[width=\linewidth]{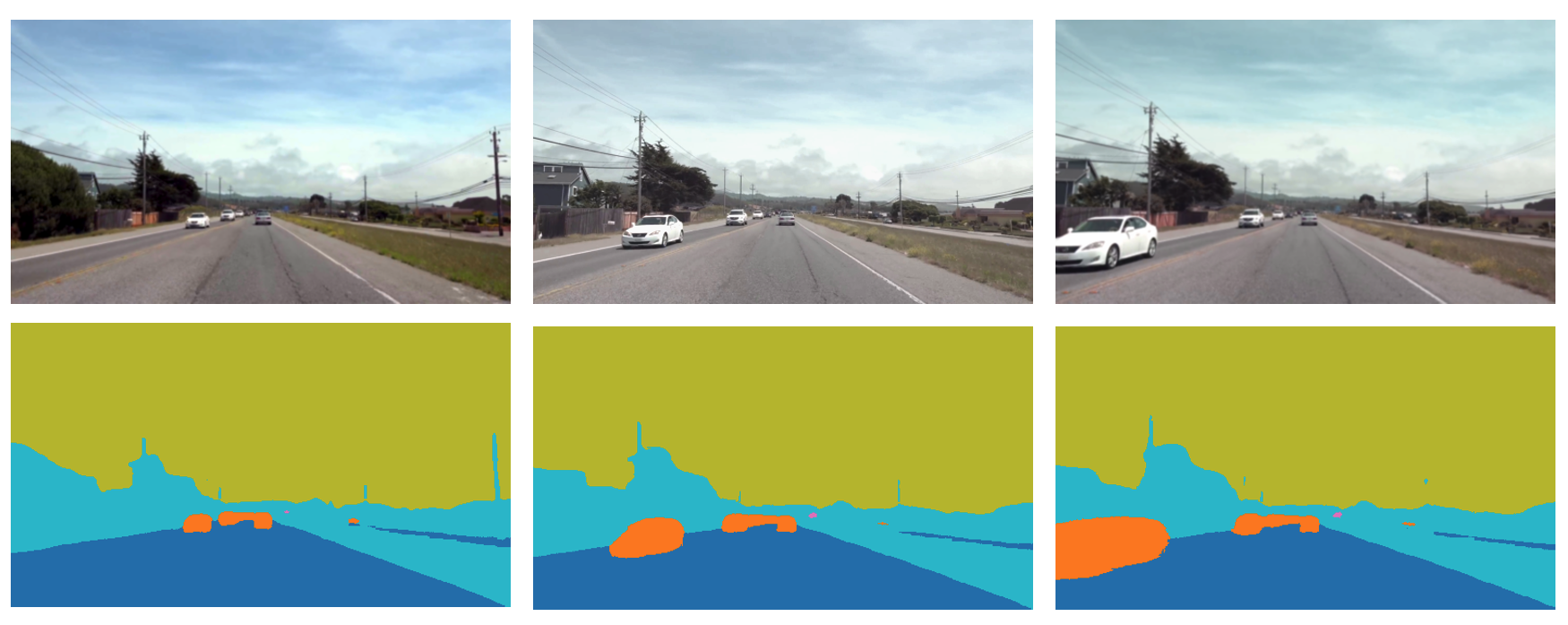}
    \caption{
    \textbf{Segmentation quality on current and future frames.}
    We show results of segmentation on the 1st frame, as well as \textbf{future} frames. LFG decouples dynamic motion from its own movement.
    }
    \label{fig:seg_qual}
\end{figure}
\subsubsection{Semantic segmentation}
\begin{table}[htbp!]
\centering
\scriptsize
\setlength{\tabcolsep}{1.5pt} 
\caption{\textbf{Semantic segmentation metrics (overall vs. predicted).}}
\vspace{2pt}

\begin{tabular}{l | A B C D | A B C D}
\toprule
\textbf{Method}
& \multicolumn{4}{c|}{\textbf{Overall}}
& \multicolumn{4}{c}{\textbf{Pred.}} \\
\cmidrule(lr){2-5} \cmidrule(lr){6-9}
& \textbf{PA} & \textbf{mIoU} & \textbf{mDice} & \textbf{FW}
& \textbf{PA} & \textbf{mIoU} & \textbf{mDice} & \textbf{FW} \\
\midrule

Static baseline
& -- & -- & -- & --
& 0.888 & 0.420 & 0.502 & 0.810 \\

SegFormer
& 0.926 & 0.677 & 0.744 & 0.723
& 0.926 & 0.680 & 0.747 & 0.725 \\

MaskFormer
& 0.922 & 0.760 & 0.829 & 0.760
& -- & -- & -- & -- \\

\textcolor{blue!60}{LFG}
& 0.947 & 0.768 & 0.827 & 0.770
& 0.942 & 0.751 & 0.814 & 0.759 \\

\bottomrule
\end{tabular}

\label{tab:semantic_results}
\vspace{-4pt}
\end{table}

We evaluate on semantic segmentation using KITTI-360~\cite{liao2022kitti} with samples of 6 consecutive frames for 200 varied sequences. We compare:
the segmentation teacher model SegFormer with all 6 RGB images as input, a MaskFormer baseline evaluated on overall frames (no future prediction), and our model with only the first 3 frames as input while predicting for all 6 frames. To measure the model's ability to anticipate future scene layout, we also provide the score between the ground truth semantic segmentation of the \textbf{third frame} compared to the following frames.
We report standard segmentation metrics (pixel accuracy, mIoU, mDice, frequency weighted IoU) on all frames and only future frames. Table~\ref{tab:semantic_results} shows that SegFormer is a stronger baseline than MaskFormer in this setting, and that our model not only beats its SegFormer teacher on overall semantic segmentation, but also on future frames where the teacher model was fed the RGB images and LFG was not.

\subsubsection{Monocular depth estimation}
\begin{table}[h]
\centering
\scriptsize   
\setlength{\tabcolsep}{1.5pt}
\caption{\textbf{Depth estimation results for overall and predicted frames.}}
\vspace{2pt}

\begin{tabular}{c l | E F | E F}
\toprule
\textbf{Dataset} & \textbf{Method} 
& \multicolumn{2}{c|}{\textbf{Overall}} 
& \multicolumn{2}{c}{\textbf{Predicted}} \\
\cmidrule(lr){3-4} \cmidrule(lr){5-6}
& & \textbf{AbsRel} & \textbf{RMSE} 
  & \textbf{AbsRel} & \textbf{RMSE} \\
\midrule

\multirow{4}{*}{KITTI-360}
& $\pi^3$  & 0.26 ± 0.08 & 4.37 ± 0.65 & 0.26 ± 0.07 & 4.37 ± 0.66 \\
& LFG  & 0.27 ± 0.07 & 4.38 ± 0.64 & 0.31 ± 0.11 & 4.38 ± 0.68 \\
& VGGT  & -- & 4.46 ± 0.82 & -- & 4.46 ± 0.82 \\
& DA3  & -- & 4.43 ± 0.81 & -- & 4.44 ± 0.81 \\
\midrule

\multirow{2}{*}{Waymo}
& $\pi^3$  & 0.19 ± 0.12 & 6.68 ± 3.10 & 0.19 ± 0.12 & 6.70 ± 3.13 \\
& LFG  & 0.21 ± 0.11 & 6.87 ± 2.72 & 0.22 ± 0.11 & 7.12 ± 2.81 \\
\bottomrule
\end{tabular}

\label{tab:depth_results}
\vspace{-4pt}
\end{table}

For the monocular depth prediction, we evaluate on the KITTI-360 and Waymo open dataset~\cite{sun2020scalability} with 200 sequences of 6 frames each. We compute root mean square error in meters after a scale and shift alignment with ground truth depth, and absolute relative depth error.  Similar to semantic segmentation, we use the sample of 6 frames and give all of them to the teacher model $\pi^3$ and the first 3 to our model. We also include strong monocular baselines (VGGT and DA3) to contextualize teacher quality; these results indicate that $\pi^3$ remains the strongest teacher in our setting. The results provided in Table~\ref{tab:depth_results} show that the depth prediction accuracy is on par with the teacher model (within 1 meter across the board) and only slightly worse on predicted future frames. More visualizations can be found in the supplementary.
\input{figures/full_pc_images}
\noindent\textbf{Point cloud reconstruction.} Fig.~\ref{fig:point_cloud_duel} provides a qualitative comparison of full point cloud reconstructions from LFG and $\pi^3$, illustrating that LFG preserves geometric structure and camera motion even when predicting future frames.

\subsubsection{Trajectory prediction}
\begin{table}[h]
\centering
\footnotesize
\setlength{\tabcolsep}{4pt}
\caption{\textbf{Trajectory estimation results.} RelPos is split into rotation (deg) and translation (m).}
\vspace{2pt}

\begin{tabular}{c l | c c c}
\toprule
\textbf{Dataset} & \textbf{Method}
& \textbf{ATE} & \textbf{Rot} & \textbf{Trans} \\
\midrule

\multirow{2}{*}{KITTI-360}
& $\pi^3$  & 0.43 & 1.32 & 0.31 \\
& LFG  & 1.00 & 2.30 & 0.31 \\
\midrule

\multirow{2}{*}{Waymo}
& $\pi^3$  & 0.02 & 0.98 & 0.44 \\
& LFG  & 0.08 & 1.00 & 0.44 \\
\bottomrule
\end{tabular}

\label{tab:traj_results}
\vspace{-4pt}
\end{table}

As our model predicts camera poses of input 3 frames and future 3 frames, we evaluate the trajectory prediction on KITTI-360 and Waymo open dataset (200 sequences of 6 frames each), and compare it to $\pi^3$ with all 6 frames as input. We report Absolute Trajectory Error (ATE), rotation error (Rot), and translation error (Trans). ATE measures the discrepancy between predicted and ground-truth trajectories after alignment. Rot and Trans denote the mean angular rotation error (deg) and mean translation error (m), respectively. In Table~\ref{tab:traj_results}, we can observe that while the metrics are slightly worse that the teacher model, the result is still competitive, considering that our model does not have access to the last 3 frames.

\begin{figure}[t]
    \centering
    \begin{tabular}{c c c c}
        & \textbf{RGB} & \textbf{LFG Motion} & \textbf{Pseudo Motion} \\

        \rotatebox{90}{~~~~~~~~~~~\textbf{Frame 1}} &
        \includegraphics[width=0.26\linewidth]{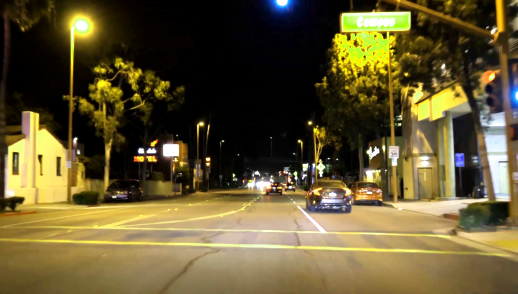} &
        \includegraphics[width=0.26\linewidth]{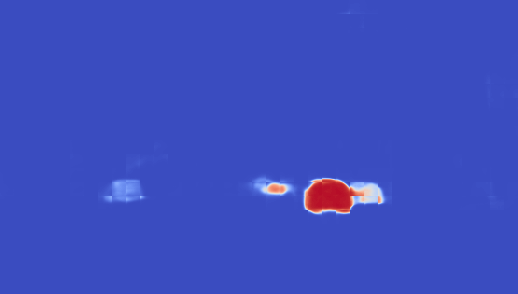} &
        \includegraphics[width=0.26\linewidth]{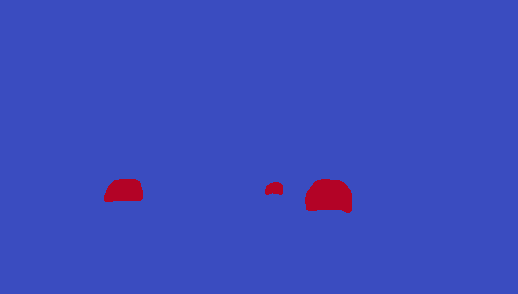} \\[3pt]

        \rotatebox{90}{~~~~~~~~~~~\textbf{Frame 2}} &
        \includegraphics[width=0.26\linewidth]{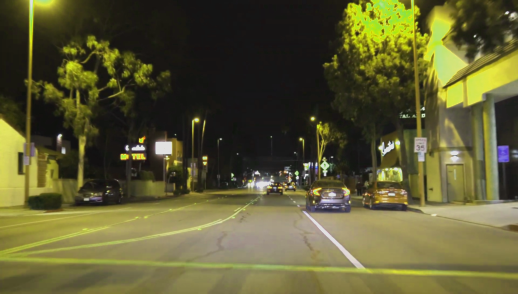} &
        \includegraphics[width=0.26\linewidth]{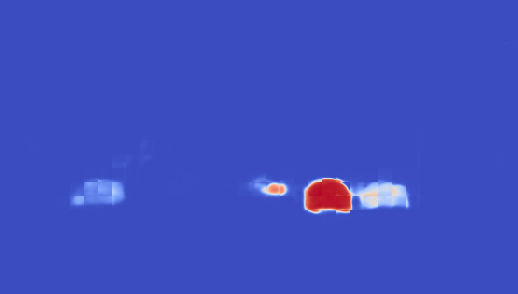} &
        \includegraphics[width=0.26\linewidth]{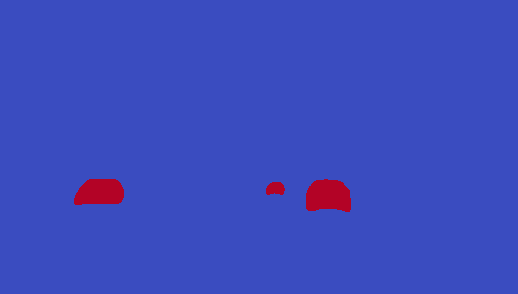} \\[3pt]

        \rotatebox{90}{~~~~~~~~~~~\textbf{Frame 3}} &
        \includegraphics[width=0.26\linewidth]{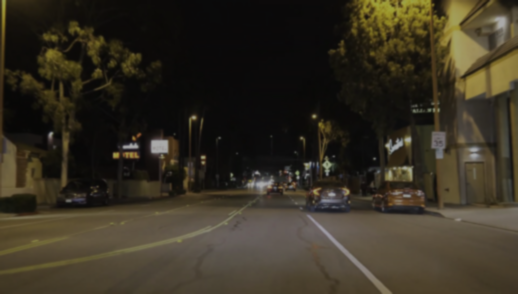} &
        \includegraphics[width=0.26\linewidth]{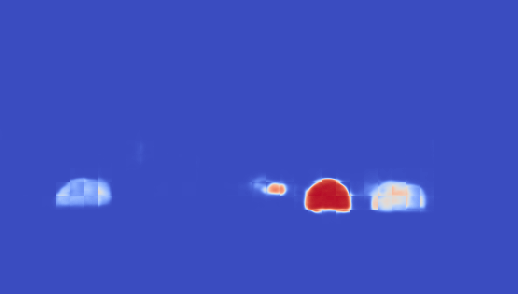} &
        \includegraphics[width=0.26\linewidth]{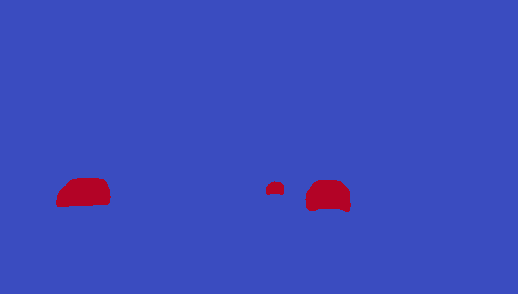} \\
    \end{tabular}

    \caption{\textbf{Failure Case of Pseudo-GT on motion}. Qualitative comparison of motion predictions (LFG vs Pseudo) with corresponding RGB frames. In this scene, the pseudo ground truth incorrectly predicts a moving car on the far left when it is parked. LFG correctly predicts the static parked car (left) and the dynamic vehicle in front of it.}
    \label{fig:motion_failure}
\end{figure}

We include a qualitative motion visualization in Fig.~\ref{fig:motion_failure}, highlighting a pseudo-ground-truth failure case where LFG correctly separates static and dynamic objects.

\subsubsection{NAVSIM planning fine-tuning}
\label{sec:exp_plan}
\begin{table}[t]
\centering
\small
\setlength{\tabcolsep}{5pt}
\renewcommand{\arraystretch}{1.15}
\caption{\textbf{Data-efficiency comparison (PDMS↑) on NAVSIM.}
LFG's pretrained encoder yields superior data efficiency, demonstrating strong performance in the low-data regime and outperforming other pretrained encoders across all label fractions.
}
\vspace{2pt}
\begin{tabular}{lcccc}
\toprule
\textbf{Method} & \textbf{Input} & \textbf{1\%} & \textbf{10\%} & \textbf{100\% Data} \\
\midrule
\rowcolor{gray!10}
DiffusionDrive    & 3Cam+L    & 64.9 & 72.6 & 88.1 \\
\midrule
DINOv3            & 1Cam & 60.0 & 75.8 & 81.4 \\
PPGeo             & 1Cam & 61.5 & 65.6 & 74.6 \\
$\pi^3$               & 1Cam & 56.2 & 77.5 & 82.8 \\
\textbf{LFG (Ours)} & 1Cam & \textbf{66.3} & \textbf{81.4} & \textbf{85.2} \\

\bottomrule
\end{tabular}
\label{tab:data_efficiency_pdms}
\vspace{-4pt}
\begin{flushleft}
\footnotesize
All pretrained encoders use a single front camera (3 frames) and the same anchor-based decoder. DiffusionDrive is trained end-to-end with a BEV-based ResNet backbone. \textbf{L} denotes LiDAR.
\end{flushleft}
\end{table}
\begin{table}[t]
\centering
\footnotesize
\setlength{\tabcolsep}{2pt}
\renewcommand{\arraystretch}{1.1}
\caption{\textbf{NAVSIM planning benchmark: single-camera LFG vs BEV-based baselines.}
Higher is better for all metrics.}
\vspace{2pt}
\begin{tabular}{lccccccc}
\toprule
\textbf{Method} & \textbf{Input} & \textbf{NC} & \textbf{DAC} & \textbf{TTC} & \textbf{C.} & \textbf{EP} & \textbf{PDMS} \\
\midrule
\multicolumn{8}{l}{\textbf{BEV Baselines}} \\
\midrule
UniAD          & 6Cam      & \underline{97.8} & 91.9 & 92.9 & 100.0 & 78.8 & 83.4 \\
TransFuser     & 3Cam+L    & 97.7             & 92.8 & 92.0 & 100.0 & 79.2 & 84.0 \\
Hydra-MDP      & 3Cam+L    & 96.9             & \underline{94.0} & 94.0 & 100.0 & 78.7 & 84.7 \\
DiffusionDrive & 3Cam+L    & 96.8             & \textbf{95.4} & \textbf{94.7} & 100.0 & \textbf{82.0} & \textbf{88.1} \\
\midrule
\textbf{LFG (Ours)} 
               & 1Cam$^{*}$  
               & \textbf{98.2}    
               & 93.7            
               & \underline{94.4} 
               & 100.0           
               & \underline{79.1}
               & \underline{85.2}
\\
\bottomrule
\end{tabular}
\label{tab:navsim_planning}
\vspace{-4pt}
\begin{flushleft}
\footnotesize
\textbf{L} = LiDAR. \quad
\textbf{1Cam$^{*}$} uses only the front-view camera with past temporal frames (3-frame input).
\end{flushleft}
\end{table}
\begin{table}[h]
\centering
\footnotesize
\setlength{\tabcolsep}{3pt}
\caption{\textbf{DiffusionDrive comparison on NAVSIM (PDMS$\uparrow$).}}
\vspace{2pt}

\begin{tabular}{l c c c c}
\toprule
\textbf{Method} & \textbf{Input} & \textbf{1\%} & \textbf{10\%} & \textbf{100\%} \\
\midrule
DiffusionDrive-DINOv2 & 3Cam+L & 57.3 & 74.4 & 81.5 \\
DiffusionDrive-DINOv2 & 1Cam & 57.5 & 73.0 & 79.7 \\
\textbf{LFG (Ours)} & \textbf{1Cam} & \textbf{66.3} & \textbf{81.4} & \textbf{85.2} \\
\bottomrule
\end{tabular}

\label{tab:diffusiondrive_r1}
\vspace{-4pt}
\end{table}

\noindent\textbf{PDMS summaries.} We report PDMS scores for NAVSIM in the data-efficiency table (Table~\ref{tab:data_efficiency_pdms}), the DiffusionDrive comparison (Table~\ref{tab:diffusiondrive_r1}), and the component/scaling ablations (Table~\ref{tab:ablations_r2}).
Across these PDMS tables, LFG is consistently strongest at 1\% and 10\% labels, and remains competitive at 100\%, outperforming DiffusionDrive-DINOv2 variants while benefitting from increased pretraining data and longer prediction horizons.
Across these PDMS tables, LFG is consistently strongest at 1\% and 10\% labels, and remains competitive at 100\%, outperforming DiffusionDrive-DINOv2 variants while benefitting from increased pretraining data and longer prediction horizons.
\textbf{Data efficiency.}
\cref{tab:data_efficiency_pdms} evaluates how well different pretrained encoders transfer to NAVSIM planning as we vary the amount of training data. Among pretrained encoders, LFG consistently achieves the best PDMS across all label fractions: at 10\% labels, LFG attains 81.4 PDMS, matching the full-data performance of DINOv3, which highlights the effectiveness of our in-the-wild video pretraining. We attribute these gains to the encoder's stronger temporal understanding of the scene, allowing it to better leverage short past frame sequences for planning. It surpasses both one of its teachers $\pi^3$ and PPGeo \cite{wu2023ppgeo}, demonstrating how both powerful feedforward architectures need semantic and temporal understanding of the future.  More ablations are provided in the supplementary.
\begin{table}[h]
\centering
\footnotesize
\setlength{\tabcolsep}{3pt}
\caption{\textbf{Component and scaling ablations on NAVSIM (PDMS$\uparrow$).}}
\vspace{2pt}

\begin{tabular}{l c c c}
\toprule
\textbf{Setting} & \textbf{1\%} & \textbf{10\%} & \textbf{100\%} \\
\midrule
\rowcolor{blue!6} Original setting & 66.3 & 81.4 & 85.2 \\
\rowcolor{green!6} {+ 2$\times$ pretraining data} & 76.6 & 82.3 & 84.8 \\
\rowcolor{orange!8} {+ Longer prediction horizon} & 80.5 & 84.4 & 84.8 \\
\rowcolor{red!6} {- Seg, Motion} & 64.8 & 77.1 & 84.6 \\
\rowcolor{purple!6} {- Autoregressive head} & 66.3 & 77.7 & 84.2 \\
\bottomrule
\end{tabular}

\label{tab:ablations_r2}
\vspace{-4pt}
\end{table}
\textbf{Ablations.}
Table~\ref{tab:ablations_r2} shows that scaling pretraining data and extending the prediction horizon both improve PDMS at low-label regimes, while removing segmentation/motion supervision or the autoregressive head degrades performance, confirming the importance of these components.

\textbf{Benchmark results.}
Compared to prior methods on NAVSIM (\cref{tab:navsim_planning}), LFG, using only single front-view camera inputs, outperforms heavily engineered BEV-based baselines such as UniAD~\cite{hu2023_uniad} and Hydra-MDP~\cite{li2024hydramdp}, which rely on multi-view cameras and/or LiDAR. LFG achieves the best Not at-fault collision (NC) score (98.2) and competitive TTC and EP scores (\underline{94.4} and \underline{79.1}), resulting in an overall PDMS of \underline{85.2}. This demonstrates that a single-camera encoder pretrained with large-scale video can rival specialized BEV-based systems that leverage significantly richer sensor suites.

\section{Conclusion}
\label{sec:conclusion}


In all, LFG learns directly from in-the-wild, unposed driving videos, and thanks to its strong pretrained encoder, it achieves competitive planning performance despite using only a single front-view camera. For fairness, we compare against DiffusionDrive-DINOv2 variants that use both multi-camera+LiDAR and single-camera inputs, and LFG remains stronger in this setting. For future direction, LFG predicts only short-term futures (3–6 frames), and extending the autoregressive module to longer or multi-scale temporal horizons may improve long-range reasoning. Second, we use only a single front-view camera, reflecting the fact that most in-the-wild driving videos provide only one viewpoint; while this setting already highlights the strength of video-based geometric priors, incorporating multi-view cues could further improve robustness in complex scenes. As larger multi-camera datasets such as the recently released PhysicalAI-Autonomous-Vehicles dataset~\cite{nvidia2025physicalai} become available, exploring multi-view training represents a promising direction for future work.




{
    \small
    \bibliographystyle{ieeenat_fullname}
    \bibliography{main}
}


\clearpage
\setcounter{page}{1}
\maketitlesupplementary
\renewcommand{\thesection}{\Alph{section}}
\setcounter{section}{0}
\renewcommand{\thefigure}{A\arabic{figure}}
\setcounter{figure}{0}

\setcounter{table}{0}
\renewcommand{\thetable}{A\arabic{table}}

This supplementary material provides additional results and implementation details.
We include full training configurations in \cref{supp:traing_details}, planning fine-tuning and baseline descriptions in \cref{supp:plan_details}, and extended qualitative visualizations: segmentation in \cref{supp:seg_vis}, motion in \cref{supp:motion_vis}, and depth in \cref{supp:depth_vis}.

\section{Training Details}\label{supp:traing_details}
For reproducibility, we share specific training details of our model. We train LFG on top of the pretrained $\pi^3$, keeping the DINOv2 encoder frozen, as well as the confidence, camera, and point decoders, including automatic mixed precision (bfloat16) to speed up training. 

To obtain motion masks from \textbf{Grounded SAM2} and \textbf{CoTracker3}, we first query \textbf{Grounded DINO} using the object priors \emph{car}, \emph{vehicle}, and \emph{person}, which yields an initial set of candidate instance masks. Each mask is then processed with \textbf{CoTracker3}, using a grid size of 80 and a motion threshold of $0.1$ in the normalized geometric space. An object is classified as dynamic if it exhibits motion in the majority of frames.

For the segmentation loss, we apply class-specific weighting across seven categories to address inherent frequency imbalances in driving scenes. Specifically, we assign weights of $0.5$ to \emph{road}, $1.2$ to \emph{vehicle}, $1.6$ to \emph{person}, $1.8$ to both \emph{traffic light} and \emph{traffic sign}, $0.3$ to \emph{sky}, and $0.2$ to \emph{background/buildings}. These weights remain fixed throughout training and were found to provide a stable and effective balance across diverse urban environments.

We apply VGGT-style photometric augmentations during training. Color jittering perturbs brightness, contrast, and saturation by $\pm 40\%$ ($0.4$) and hue by $\pm 10\%$ ($0.1$). Random grayscale is applied with probability $0.1$. Additionally, we apply random Gaussian blur with probability $0.2$, using a sigma sampled uniformly from $[0.1, 2.0]$. We resize all images to $(294, 518)$, and train on the prior 3 images, predicting outputs for the next 3 images, but additionally train the motion head (final stage) on both the 3 prior images and 6 prior images. We vary the time between each image, randomly sampling from $2, 5, 10$Hz.

\section{Planning Fine-tuning and Baseline Details
}\label{supp:plan_details}
We fine-tune all models on the NAVSIM planning benchmark using only the front-view camera over three consecutive frames to predict 4s future ego trajectories. Unless otherwise specified, all baseline vision encoders are kept frozen and we only train lightweight causal attention adapters and a shared anchor-based trajectory decoder.
\paragraph{Common planning head.}
For all methods (ours and baselines), we employ the same anchor-based trajectory decoder. Following DiffusionDrive~\cite{liao2025diffusiondrive}, we adopt $K=20$ trajectory anchors obtained by K-means clustering over ground-truth futures; however we omit the diffusion component and any iterative refinement to keep the architecture simple. After causal temporal aggregation from vision encoder's embedding, the decoder attends over trajectory anchors and across modes, and in a single forward pass predicts (i) confidence scores for each of the $K$ modes and (ii) coordinate offsets for each waypoint along each mode. At test time, the highest-confidence mode is selected as the final plan. All models predict 8 waypoints at 0.5s intervals (a 4s horizon) and are trained with a combination of focal loss (classification over modes) and L1 regression loss on waypoints.

\paragraph{Temporal aggregation.}
For each front-view frame, the pretrained encoder produces high-dimensional autonomy tokens encoding ego motion and scene context. To exploit temporal structure, we apply a small causal self-attention module across the three input frames’ embeddings. The resulting aggregated features are passed into the trajectory decoder. With our method (LFG), since the encoder has already been pretrained with temporal reasoning, we use the last set of future autonomy tokens directly, which provides a temporally consistent representation for the planning head to condition on.

\paragraph{Baselines and training protocol.}
We evaluate three frozen encoders: PPGeo (geometric pre-training), DINOv3 (self-supervised ViT), and Pi3 (4D self-supervised learning). Each is followed by the same causal temporal adapter and shared anchor-based planning head. Our method (LFG) uses the pretrained temporal autoregressive encoder described in the main paper (also kept frozen), along with a lightweight multi-modal trajectory decoder. All models are optimized with AdamW and a cosine learning-rate schedule, and are trained under identical data-scaling regimes using 1\%, 10\%, and 100\% of NAVSIM training data with learning rate 1e-4 to study data efficiency.

For {DiffusionDrive}, we follow the publicly released implementation (code available on GitHub) which uses three front-view cameras plus LiDAR input and their corresponding hyper parameters. 
\begin{table}[t]
\centering
\small
\setlength{\tabcolsep}{5pt}
\renewcommand{\arraystretch}{1.15}
\caption{\textbf{Comparing PPGeo with different pretraining data-source.}
PPGeo$^\ast$ indicates that the model is pretrained on the \emph{same OpenDV dataset} used by LFG.}
\vspace{2pt}
\begin{tabular}{lcccc}
\toprule
\textbf{Method} & \textbf{Input} & \textbf{1\%} & \textbf{10\%} & \textbf{100\% Data} \\
\midrule
PPGeo & 1Cam & 61.5 & 65.6 & 74.6 \\
PPGeo$^\ast$ & 1Cam & 59.8 & 70.0 & 76.4 \\
\textbf{LFG (Ours)} & 1Cam & \textbf{66.3} & \textbf{81.4} & \textbf{85.2} \\

\bottomrule
\end{tabular}
\label{tab:sup_data_efficiency_ppgeo}
\vspace{-4pt}
\begin{flushleft}
\footnotesize
\end{flushleft}
\end{table}

For PPGeo, in \cref{tab:navsim_planning} we use the publicly released ResNet-34 encoder from the PPGeo repository\footnote{\url{https://github.com/OpenDriveLab/PPGeo}} pretrained with geometric self-supervision~\cite{wu2023ppgeo}. The original PPGeo encoder is pretrained using the YouTube driving video dataset introduced in the ACO project\footnote{\url{https://github.com/metadriverse/ACO}}. To isolate the impact of pre-training data source, we evaluate a variant, PPGeo$^\ast$, where we replicate the same geometric pre-training procedure but restrict the pre-training corpus to exactly the data used by LFG. As shown in \cref{tab:sup_data_efficiency_ppgeo}, PPGeo$^\ast$ slightly improves performance at higher label fractions but still under-performs LFG by a wide margin, highlighting that LFG’s 4-D temporal pre-training paradigm provides inductive biases that align more directly with downstream planning.

\paragraph{NAVSIM metrics}
The NAVSIM benchmark uses a composite score called the Predictive Driver Model Score (PDMS) to evaluate planning performance. PDMS is computed in two phases:  
(i) two hard-multiplier subscores  \textbf{No at-fault Collisions (NC)} and \textbf{Drivable Area Compliance (DAC)}  that immediately zero the scenario score if violated;  
(ii) a weighted average of three performance subscores  \textbf{Ego Progress (EP)}, \textbf{Time-to-Collision (TTC)}, and \textbf{Comfort (C)} — reflecting route progress, safety margin, and motion smoothness.

Formally,:
\[
\text{PDMS} = \bigl(\text{NC} \times \text{DAC}\bigr)\times \frac{5\,\text{EP} + 5\,\text{TTC} + 2\,C}{5 + 5 + 2}
\]
Here:
\begin{itemize}
  \item NC = 1 if no at-fault collision, = 0.5 if a collision with a static object, = 0 otherwise.
  \item DAC = 1 if the ego vehicle remains within the drivable area for the entire rollout, = 0 if it leaves.
  \item EP is the ratio of actual route progress achieved to a safe upper bound (clipped to [0,1]).
  \item TTC = 1 if the minimum time-to-collision along the 4s horizon exceeds a fixed threshold, else = 0.
  \item C = 1 if all vehicle kinematic thresholds (acceleration, jerk) remain within comfort bounds, else = 0.
\end{itemize}

All metrics are evaluated via a non-reactive 4-second rollout in the benchmark simulator in the test set ~12k samples.

\section{Segmentation Visualizations}\label{supp:seg_vis}
\begin{figure*}[t]
    \centering
    \begin{tabular}{c c c c}
        & \textbf{RGB} & \textbf{LFG Semantics} & \textbf{SegFormer Semantics} \\

        \rotatebox{90}{~~~~~~~~~~~\textbf{\textcolor{blue}{Frame 1}}} &
        \includegraphics[width=0.30\linewidth]{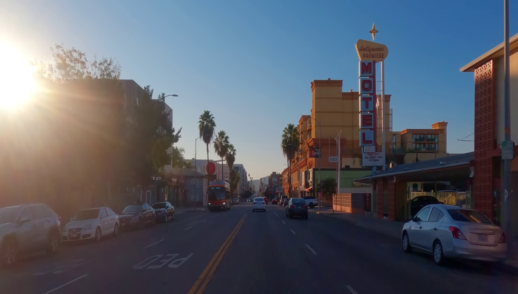} &
        \includegraphics[width=0.30\linewidth]{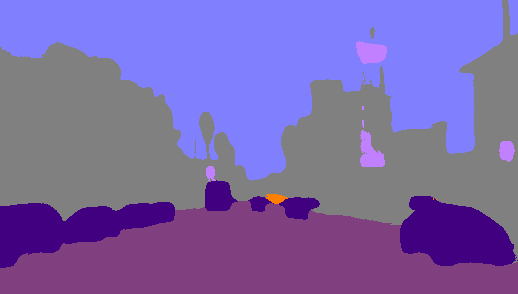} &
        \includegraphics[width=0.30\linewidth]{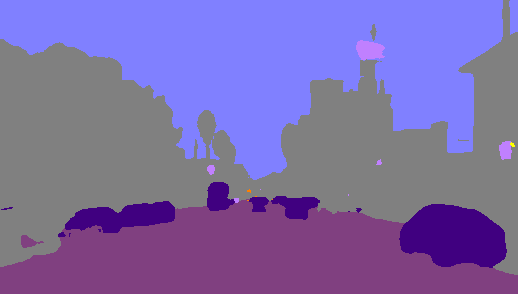} \\[4pt]

        \rotatebox{90}{~~~~~~~~~~~\textbf{\textcolor{blue}{Frame 2}}} &
        \includegraphics[width=0.30\linewidth]{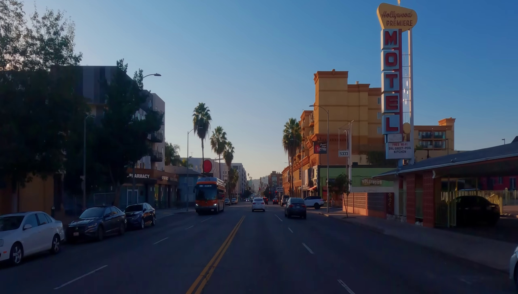} &
        \includegraphics[width=0.30\linewidth]{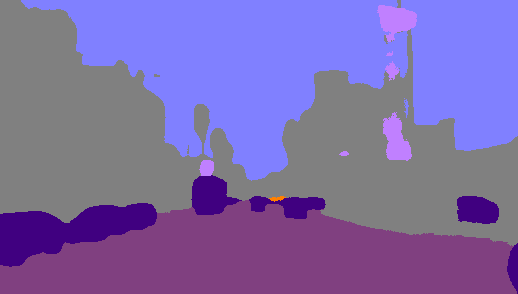} &
        \includegraphics[width=0.30\linewidth]{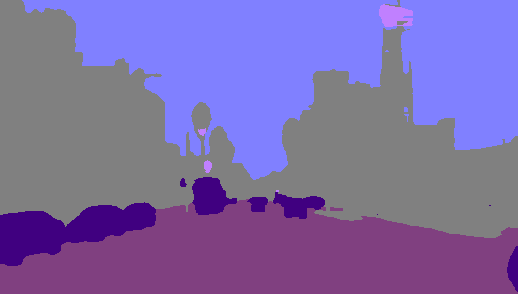} \\[4pt]

        \rotatebox{90}{~~~~~~~~~~~\textbf{\textcolor{red}{Frame 4}}} &
        \includegraphics[width=0.30\linewidth]{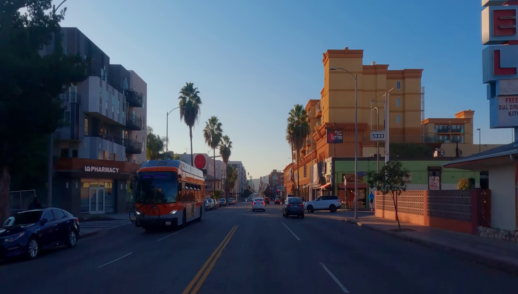} &
        \begin{tikzpicture}
          \node[inner sep=0pt] (img)
            {\includegraphics[width=0.30\linewidth]{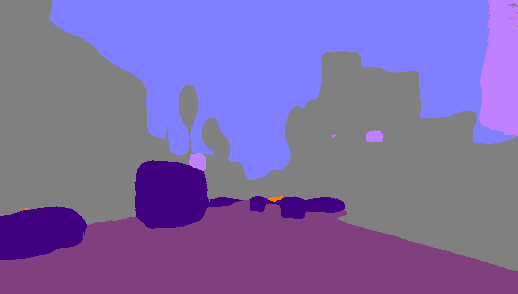}};
          \draw[red, dashed, thick] (img.south west) rectangle (img.north east);
        \end{tikzpicture} &
        \includegraphics[width=0.30\linewidth]{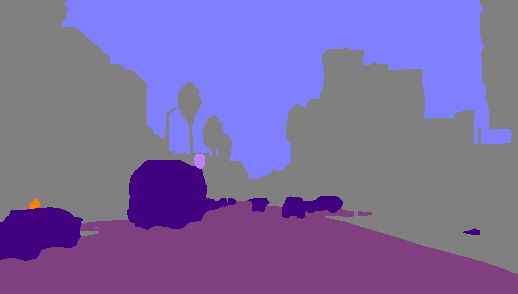}
        \\[4pt]

        \rotatebox{90}{~~~~~~~~~~~\textbf{\textcolor{red}{Frame 6}}} &
        \includegraphics[width=0.30\linewidth]{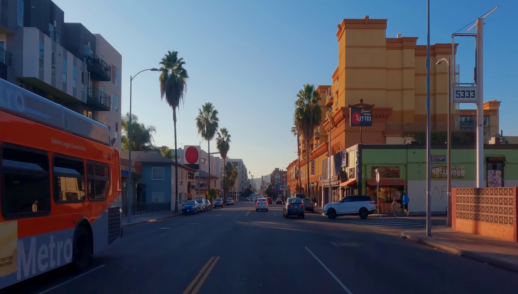} &
        \begin{tikzpicture}
          \node[inner sep=0pt] (img)
            {\includegraphics[width=0.30\linewidth]{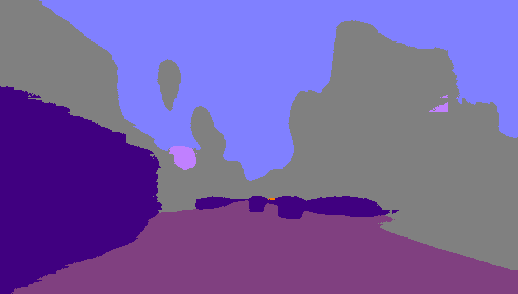}};
          \draw[red, dashed, thick] (img.south west) rectangle (img.north east);
        \end{tikzpicture} &
        \includegraphics[width=0.30\linewidth]{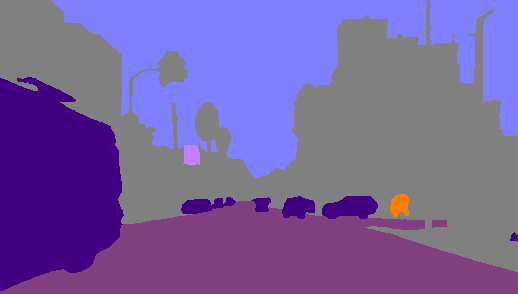}
        \\[4pt]\\
    \end{tabular}

    \caption{Qualitative comparison of semantic segmentation across RGB, LFG, and SegFormer for 
    \textcolor{blue}{current frames~1 and~2} (with ground-truth input) and 
    \textcolor{red}{future frames~4 and~6}. 
    Dashed red outlines denote predicted frames with \emph{no ground-truth image input}, produced solely from the model’s future tokens.}
    \label{fig:semantic_comparison_1}
\end{figure*}

We show segmentation visualizations on the OpenDV dataset, with sample unposed images, and the teacher SegFormer model outputs as in Fig. \ref{fig:semantic_comparison_1}, on a $5$hz scene. We find that LFG performs very competitively with its SegFormer teacher on the current frames, and future predicts the motion of the moving bus as it is about to pass the ego vehicle. LFG, however, suffers from a smoothing effect in the later frames. We posit that training LFG on more steps and the entire OpenDV dataset will improve this, as well as an edge aware point map loss to improve crispness of future frame predictions.

\section{Motion Visualizations}\label{supp:motion_vis}

\begin{figure*}[t]
    \centering
    \begin{tabular}{c c c c}
        & \textbf{RGB} & \textbf{LFG Motion} & \textbf{Pseudo Motion} \\

        \rotatebox{90}{~~~~~~~~~~~\textbf{Frame 1}} &
        \includegraphics[width=0.30\linewidth]{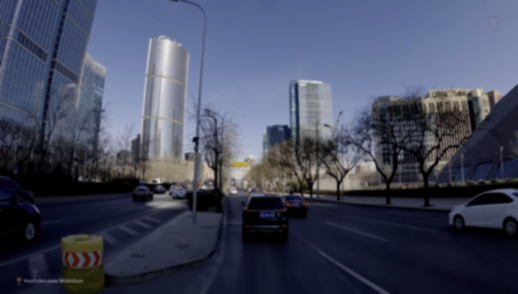} &
        \includegraphics[width=0.30\linewidth]{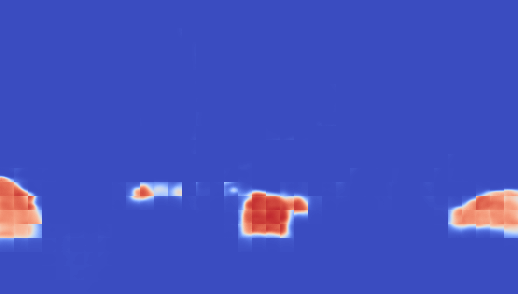} &
        \includegraphics[width=0.30\linewidth]{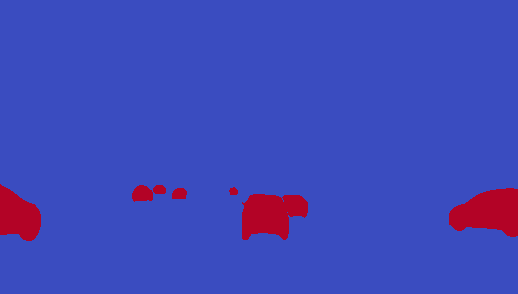} \\[4pt]

        \rotatebox{90}{~~~~~~~~~~~\textbf{Frame 2}} &
        \includegraphics[width=0.30\linewidth]{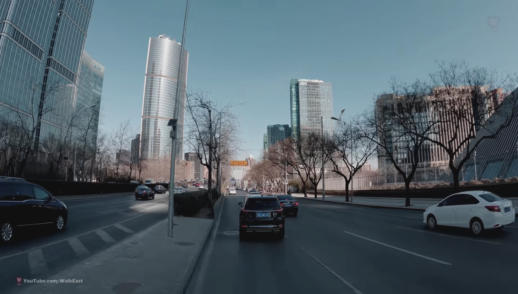} &
        \includegraphics[width=0.30\linewidth]{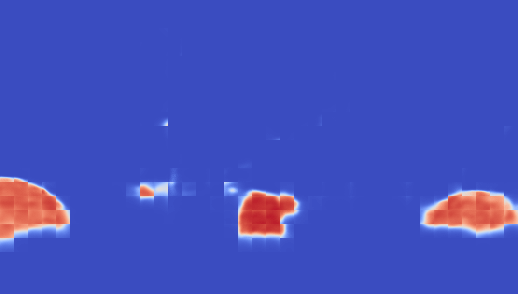} &
        \includegraphics[width=0.30\linewidth]{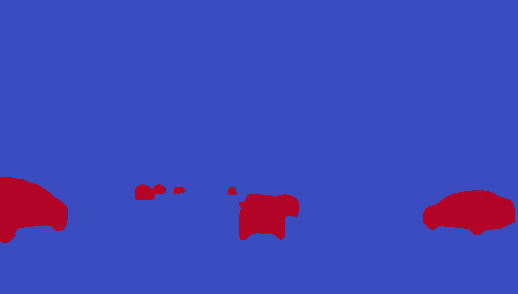} \\[4pt]

        \rotatebox{90}{~~~~~~~~~~~\textbf{Frame 3}} &
        \includegraphics[width=0.30\linewidth]{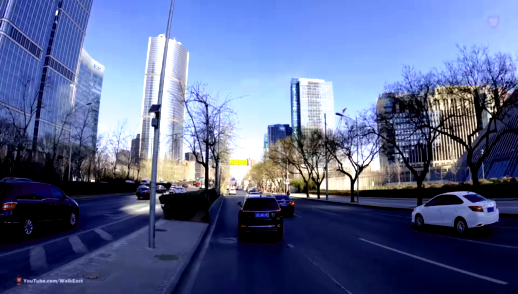} &
        \includegraphics[width=0.30\linewidth]{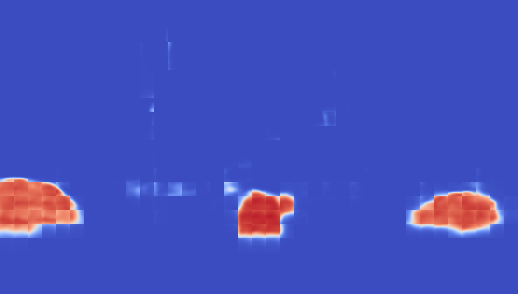} &
        \includegraphics[width=0.30\linewidth]{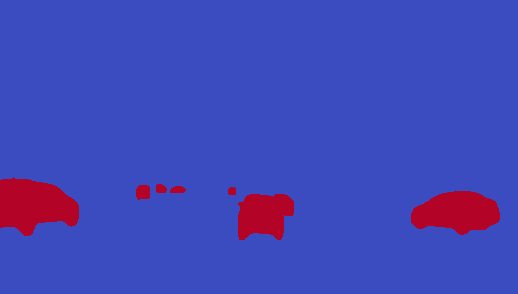} \\
    \end{tabular}

    \caption{Qualitative comparison of motion predictions (LFG vs Pseudo GT motion) with corresponding RGB frames. We show results on non-future frames to demonstrate the motion map precision on a few images. The ego vehicle is moving on a road with three nearby vehicles moving.}\
    \label{fig:motion_first}
\end{figure*}

We demonstrate motion visualization on the OpenDV dataset, seen in Fig \ref{fig:motion_first}, with current frames to emphasize the performance trained from pseudo ground truth data, on $10$Hz, but we show $3$ frames spaced apart every other frame. LFG correctly predicts the moving cars in frame from only 2D images, with a small amount of frames. Future work entails demonstrating LFG's performance for constructing dynamic Gaussian Splats, where the motion masks can be freely obtained.

\section{Depth Visualizations}\label{supp:depth_vis}

\begin{figure*}[t]
    \centering
    \begin{tabular}{c c c c}
        & \textbf{RGB} & \textbf{LFG Depth (3 Frames)} & \textbf{ $\pi^3$ Depth (All Frames)} \\

        \rotatebox{90}{~~~~~~~~~~~\textcolor{blue}{\textbf{Frame 1}}} &
        \includegraphics[width=0.30\linewidth]{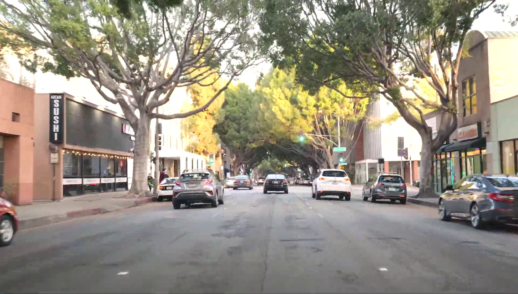} &
        \includegraphics[width=0.30\linewidth]{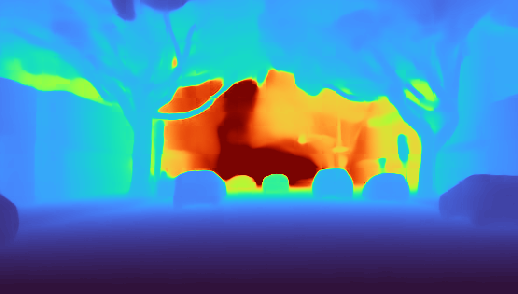} &
        \includegraphics[width=0.30\linewidth]{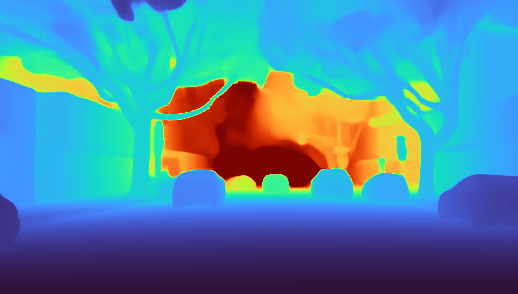} \\[4pt]

        \rotatebox{90}{~~~~~~~~~~~\textcolor{blue}{\textbf{Frame 2}}} &
        \includegraphics[width=0.30\linewidth]{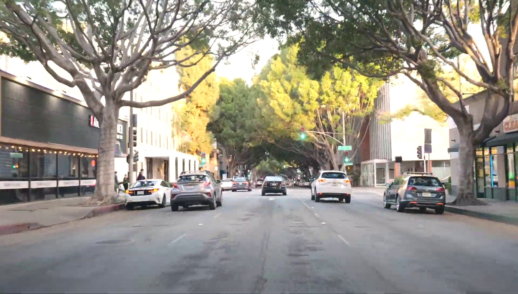} &
        \includegraphics[width=0.30\linewidth]{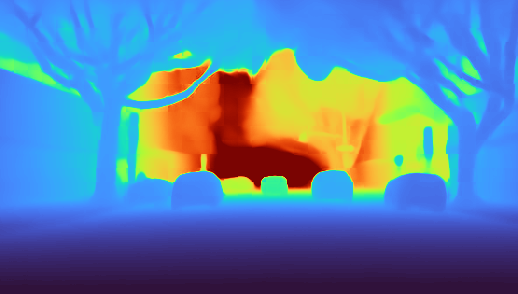} &
        \includegraphics[width=0.30\linewidth]{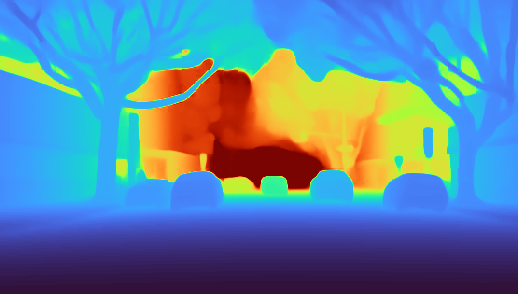} \\[4pt]

        \rotatebox{90}{~~~~~~~~~~~\textcolor{blue}{\textbf{Frame 3}}} &
        \includegraphics[width=0.30\linewidth]{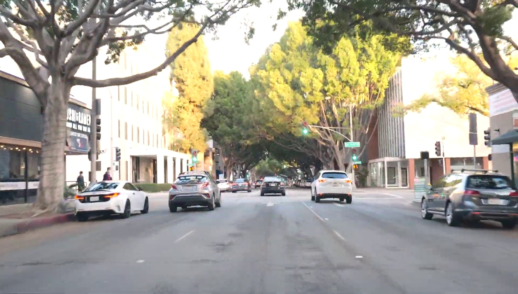} &
        \includegraphics[width=0.30\linewidth]{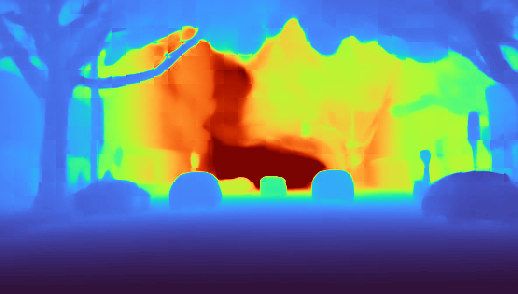} &
        \includegraphics[width=0.30\linewidth]{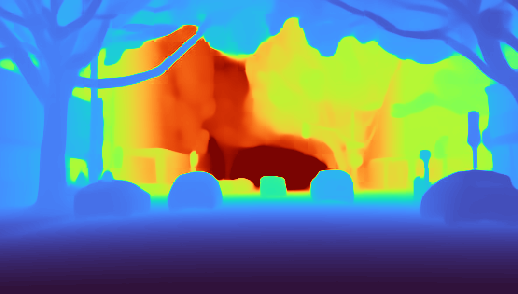} \\[4pt]

        \rotatebox{90}{~~~~~~~~~~~\textcolor{red}{\textbf{Frame 4}}} &
        \includegraphics[width=0.30\linewidth]{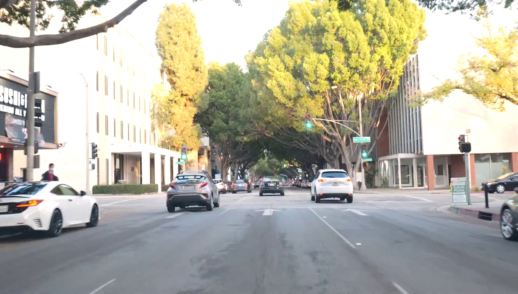} &
        \begin{tikzpicture}
          \node[inner sep=0pt] (img)
        {\includegraphics[width=0.30\linewidth]{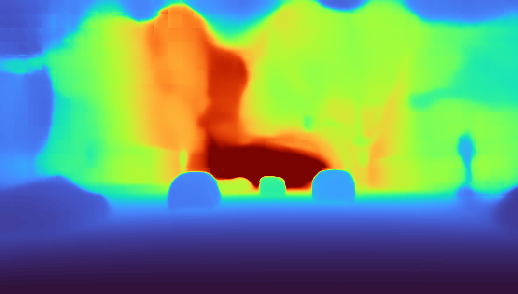}};
          \draw[red, dashed, thick] (img.south west) rectangle (img.north east);
        \end{tikzpicture}  &
        \includegraphics[width=0.30\linewidth]{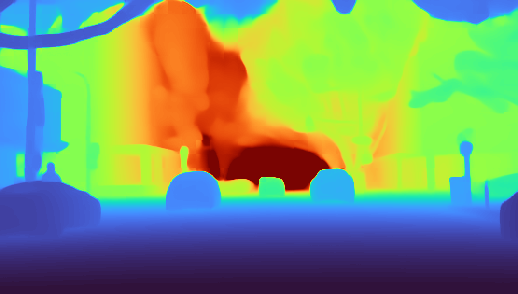} \\[4pt]

        \rotatebox{90}{~~~~~~~~~~~\textcolor{red}{\textbf{Frame 5}}} &
        \includegraphics[width=0.30\linewidth]{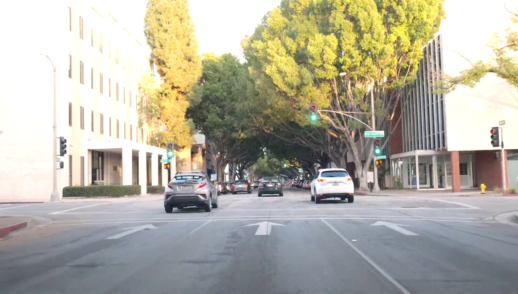} &
        \begin{tikzpicture}
          \node[inner sep=0pt] (img)
        {\includegraphics[width=0.30\linewidth]{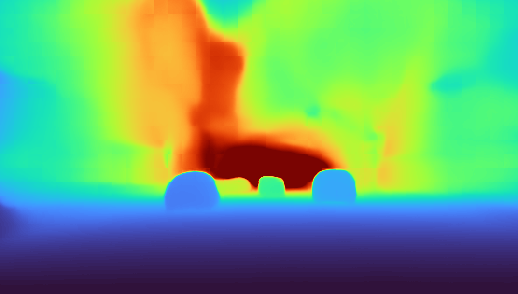}};
          \draw[red, dashed, ultra thick] (img.south west) rectangle (img.north east);
        \end{tikzpicture} &
        \includegraphics[width=0.30\linewidth]{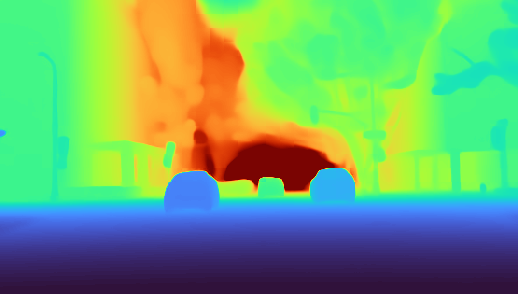} \\[4pt]

        \rotatebox{90}{~~~~~~~~~~~\textcolor{red}{\textbf{Frame 6}}} &
        \includegraphics[width=0.30\linewidth]{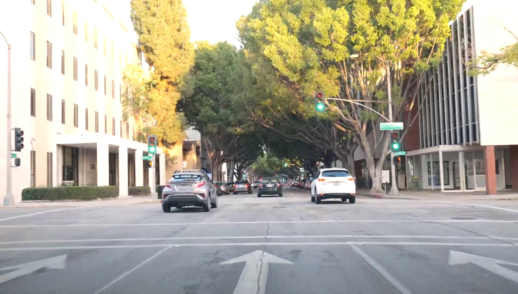} &
        \begin{tikzpicture}
          \node[inner sep=0pt] (img)
            {\includegraphics[width=0.30\linewidth]{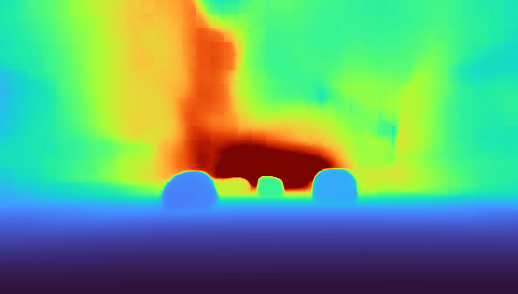}};
          \draw[red, dashed, ultra thick] (img.south west) rectangle (img.north east);
        \end{tikzpicture}
         &
        \includegraphics[width=0.30\linewidth]{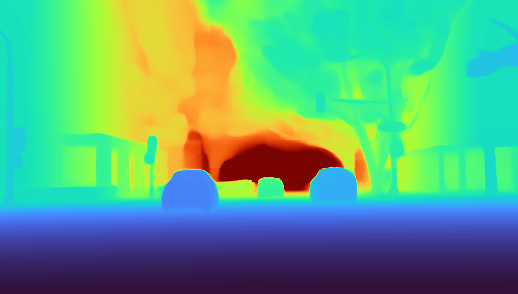} \\
    \end{tabular}

    \caption{Qualitative comparison of depth prediction for six frames (first three frames are \textcolor{blue}{blue}, the future three frames are \textcolor{red}{red}. LFG is able to decouple static and dynamic objects as it continues along the road, and future work will improve the sharpness of the last frames' predictions.
    \textbf{Dashed red} outlines denote predicted frames with \emph{no ground-truth image input}, produced solely from the model’s future tokens.}
\label{fig:depth_pi3_duel}
\end{figure*}

We show  depth visualizations of LFG compared to $\pi^3$ on validation images on our dataset, at a frequency of $5$Hz, on Fig. \ref{fig:depth_pi3_duel}. LFG performs comparable to $\pi^3$ on the seen frames, and while sharp edges are slowly lost in the future frames, LFG is able to understand dynamic and static objects, and the relative positioning of the other vehicles over time. Future work will crisp the point maps, and more results, including motion and semantic results, which are shown at the end of the supplementary.

\begin{figure*}[t]
    \centering
    \begin{tabular}{c c c c}
        & \textbf{RGB} & \textbf{LFG Semantics} & \textbf{SegFormer Semantics} \\

        \rotatebox{90}{~~~~~~~~~~~\textbf{\textcolor{blue}{Frame 1}}} &
        \includegraphics[width=0.30\linewidth]{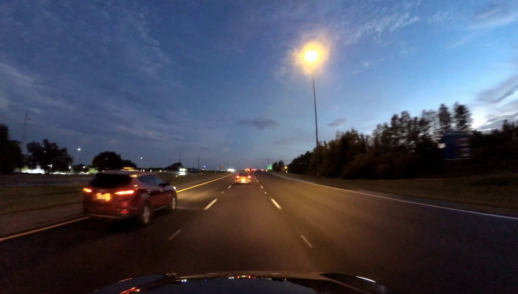} &
        \includegraphics[width=0.30\linewidth]{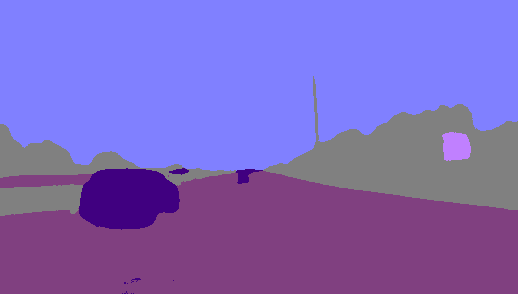} &
        \includegraphics[width=0.30\linewidth]{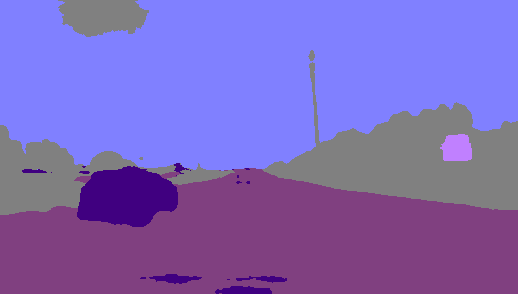} \\[4pt]

        \rotatebox{90}{~~~~~~~~~~~\textbf{\textcolor{blue}{Frame 2}}} &
        \includegraphics[width=0.30\linewidth]{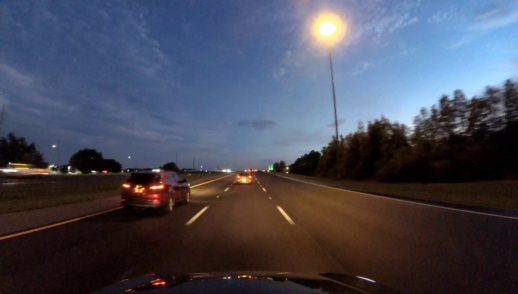} &
        \includegraphics[width=0.30\linewidth]{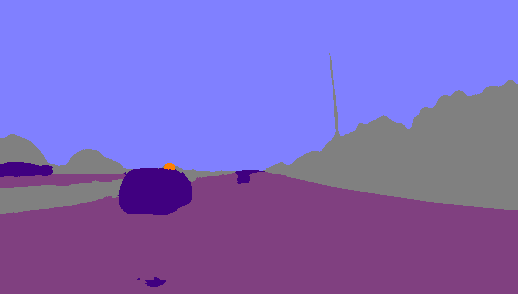} &
        \includegraphics[width=0.30\linewidth]{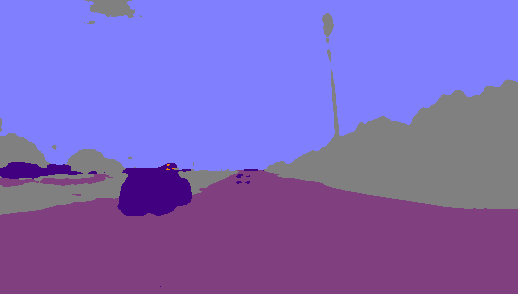} \\[4pt]

        \rotatebox{90}{~~~~~~~~~~~\textbf{\textcolor{red}{Frame 4}}} &
        \includegraphics[width=0.30\linewidth]{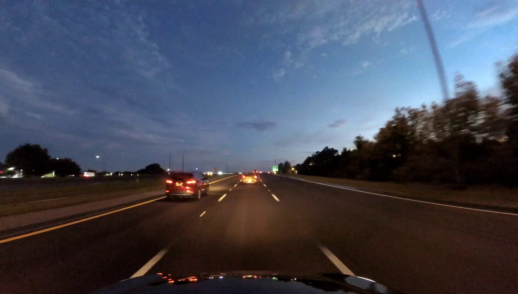} &
        \begin{tikzpicture}
          \node[inner sep=0pt] (img)
            {\includegraphics[width=0.30\linewidth]{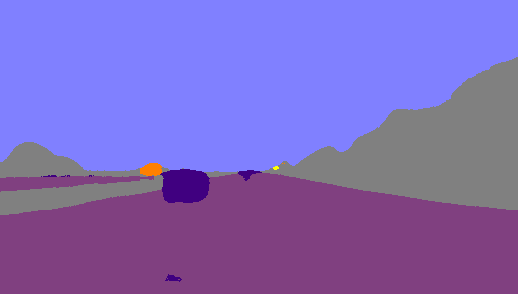}};
          \draw[red, dashed, ultra thick] (img.south west) rectangle (img.north east);
        \end{tikzpicture} &
        \includegraphics[width=0.30\linewidth]{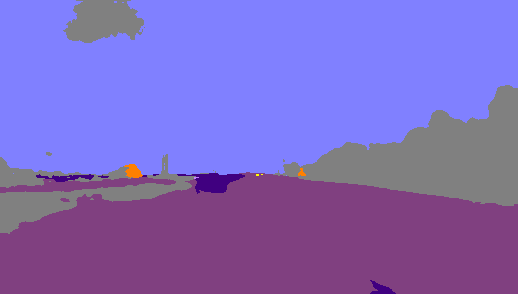}
        \\[4pt]

        \rotatebox{90}{~~~~~~~~~~~\textbf{\textcolor{red}{Frame 6}}} &
        \includegraphics[width=0.30\linewidth]{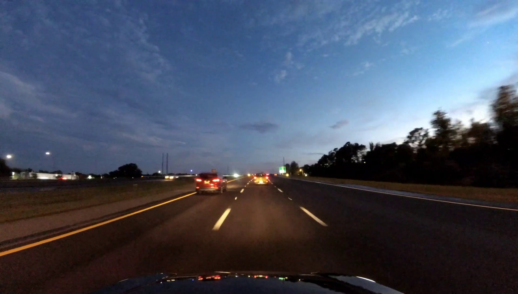} &
        \begin{tikzpicture}
          \node[inner sep=0pt] (img)
            {\includegraphics[width=0.30\linewidth]{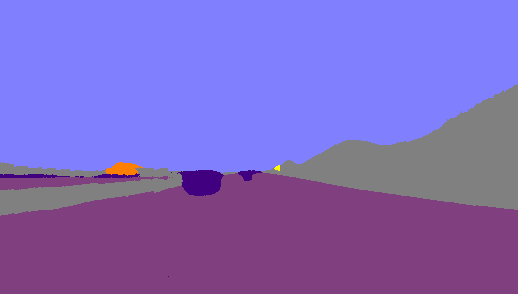}};
          \draw[red, dashed, ultra thick] (img.south west) rectangle (img.north east);
        \end{tikzpicture} &
        \includegraphics[width=0.30\linewidth]{supplementary_figures/semantics/scene4/segformer/frame_05_seg_colored.png}
        \\[4pt]\\
    \end{tabular}

    \caption{Additional qualitative comparison of semantic segmentation across RGB, LFG, and SegFormer for 
    \textcolor{blue}{current frames~1 and~2} (with ground-truth input) and 
    \textcolor{red}{future frames~4 and~6}. 
    \textbf{Dashed red} outlines denote predicted frames with \emph{no ground-truth image input}, produced solely from the model’s future tokens. LFG on current frames enjoys crisper predictions even than its teacher. }
    \label{fig:semantic_comparison_1}
\end{figure*}

\begin{figure*}[t]
    \centering
    \begin{tabular}{c c c c}
        & \textbf{RGB} & \textbf{LFG Semantics} & \textbf{SegFormer Semantics} \\

        \rotatebox{90}{~~~~~~~~~~~\textbf{\textcolor{blue}{Frame 1}}} &
        \includegraphics[width=0.30\linewidth]{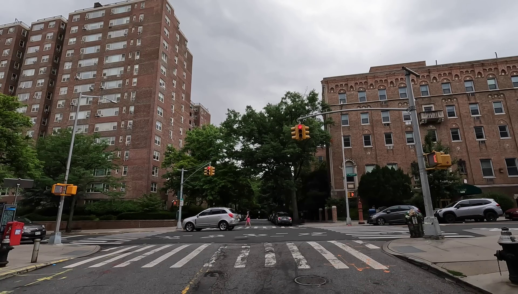} &
        \includegraphics[width=0.30\linewidth]{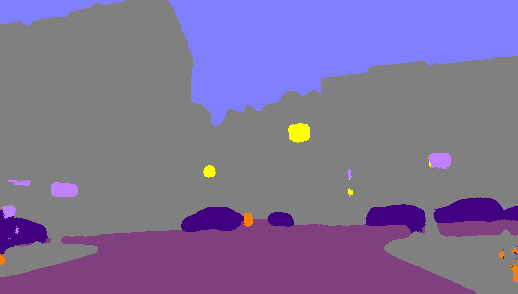} &
        \includegraphics[width=0.30\linewidth]{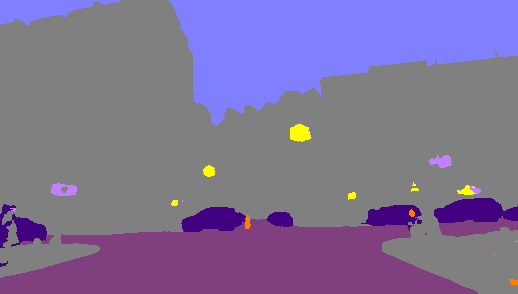} \\[4pt]

        \rotatebox{90}{~~~~~~~~~~~\textbf{\textcolor{blue}{Frame 2}}} &
        \includegraphics[width=0.30\linewidth]{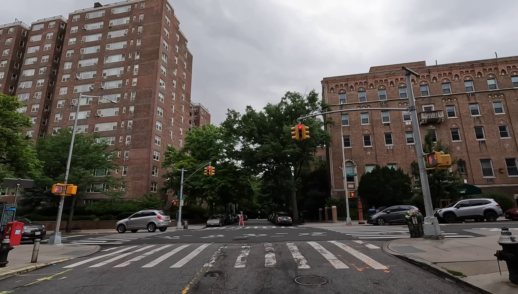} &
        \includegraphics[width=0.30\linewidth]{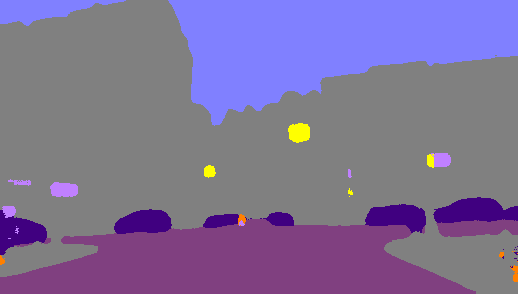} &
        \includegraphics[width=0.30\linewidth]{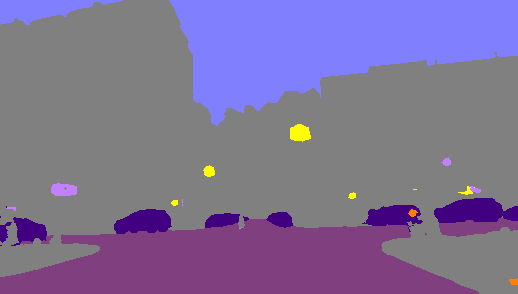} \\[4pt]

        \rotatebox{90}{~~~~~~~~~~~\textbf{\textcolor{red}{Frame 4}}} &
        \includegraphics[width=0.30\linewidth]{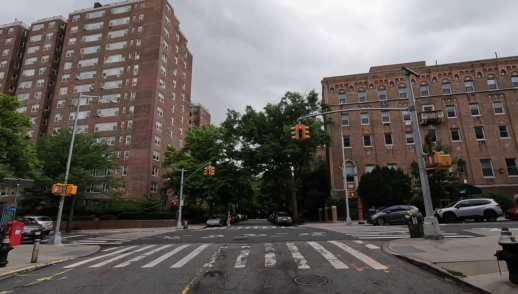} &
        \begin{tikzpicture}
          \node[inner sep=0pt] (img)
            {\includegraphics[width=0.30\linewidth]{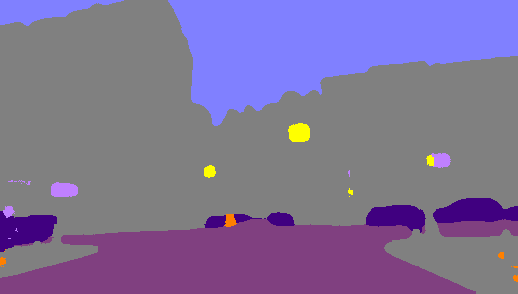}};
          \draw[red, dashed, ultra thick] (img.south west) rectangle (img.north east);
        \end{tikzpicture} &
        \includegraphics[width=0.30\linewidth]{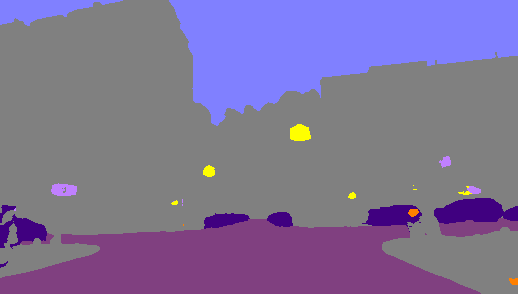}
        \\[4pt]

        \rotatebox{90}{~~~~~~~~~~~\textbf{\textcolor{red}{Frame 6}}} &
        \includegraphics[width=0.30\linewidth]{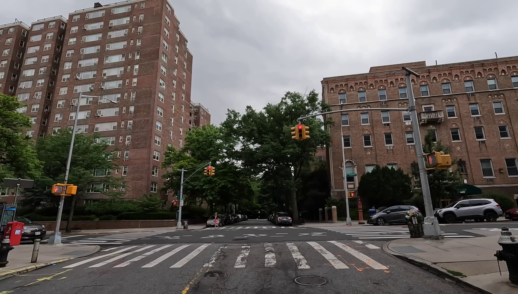} &
        \begin{tikzpicture}
          \node[inner sep=0pt] (img)
            {\includegraphics[width=0.30\linewidth]{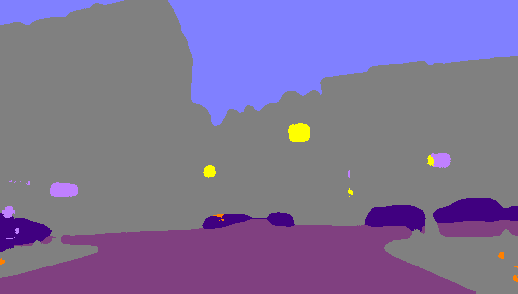}};
          \draw[red, dashed, ultra thick] (img.south west) rectangle (img.north east);
        \end{tikzpicture} &
        \includegraphics[width=0.30\linewidth]{supplementary_figures/semantics/scene5/segformer/frame_05_seg_colored.png}
        \\[4pt]\\
    \end{tabular}

    \caption{Additional qualitative comparison of semantic segmentation across RGB, LFG, and SegFormer for 
    \textcolor{blue}{current frames~1 and~2} (with ground-truth input) and 
    \textcolor{red}{future frames~4 and~6}. 
    \textbf{Dashed red} outlines denote predicted frames with \emph{no ground-truth image input}, produced solely from the model’s future tokens. LFG retains accurate predictions of cars, road, buildings, sky, traffic lights and signs, and even a person.}
    \label{fig:semantic_comparison_1}
\end{figure*}

\begin{figure*}[t]
    \centering
    \begin{tabular}{c c c c}
        & \textbf{RGB} & \textbf{LFG Motion} & \textbf{Pseudo Motion} \\

        \rotatebox{90}{~~~~~~~~~~~\textbf{Frame 1}} &
        \includegraphics[width=0.30\linewidth]{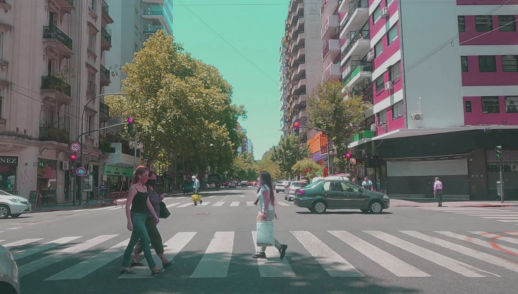} &
        \includegraphics[width=0.30\linewidth]{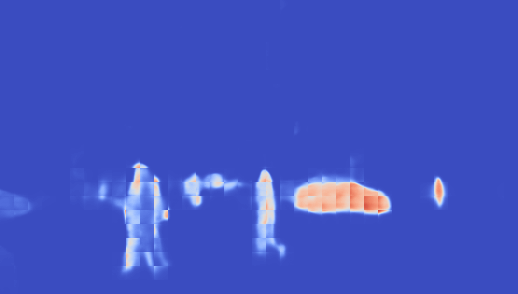} &
        \includegraphics[width=0.30\linewidth]{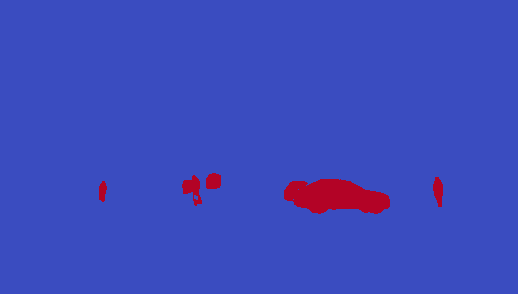} \\[4pt]

        \rotatebox{90}{~~~~~~~~~~~\textbf{Frame 2}} &
        \includegraphics[width=0.30\linewidth]{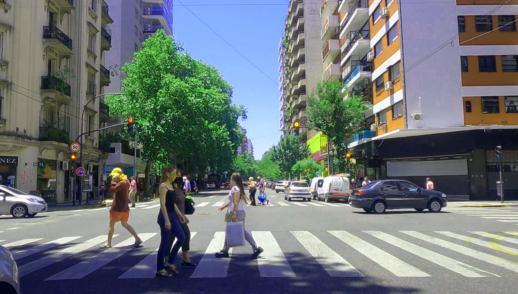} &
        \includegraphics[width=0.30\linewidth]{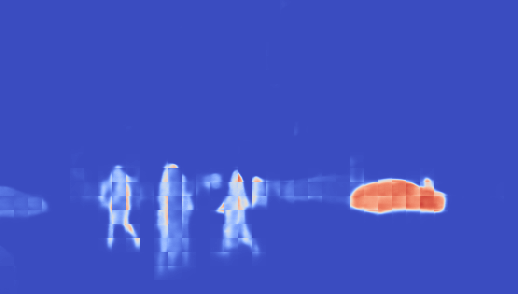} &
        \includegraphics[width=0.30\linewidth]{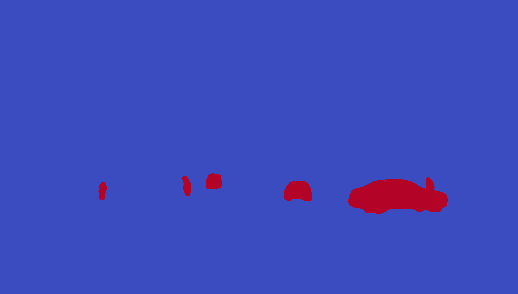} \\[4pt]

        \rotatebox{90}{~~~~~~~~~~~\textbf{Frame 3}} &
        \includegraphics[width=0.30\linewidth]{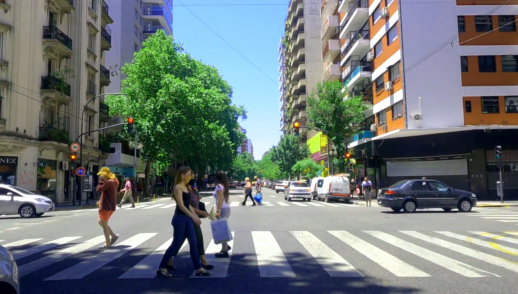} &
        \includegraphics[width=0.30\linewidth]{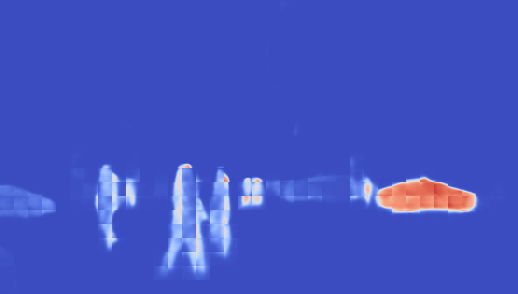} &
        \includegraphics[width=0.30\linewidth]{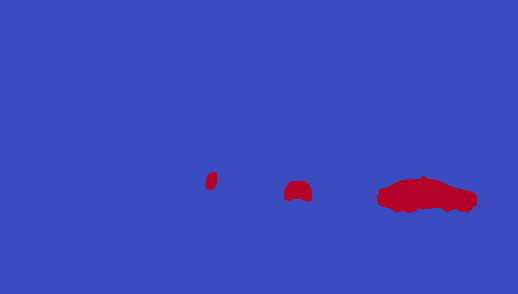} \\
    \end{tabular}

    \caption{{More qualitative comparison of motion predictions (LFG vs Pseudo) with corresponding RGB frames. In this scene, LFG predicts the moving car across the intersection, but also the close pedestrians, demonstrating that the pretrained point decoders of $\pi^3$ improve the predictions.}}
\end{figure*}

\begin{figure*}[t]
    \centering
    \begin{tabular}{c c c c}
        & \textbf{RGB} & \textbf{LFG Depth (3 Frames)} & \textbf{ $\pi^3$ Depth (All Frames)} \\

        \rotatebox{90}{~~~~~~~~~~~\textcolor{blue}{\textbf{Frame 1}}} &
        \includegraphics[width=0.30\linewidth]{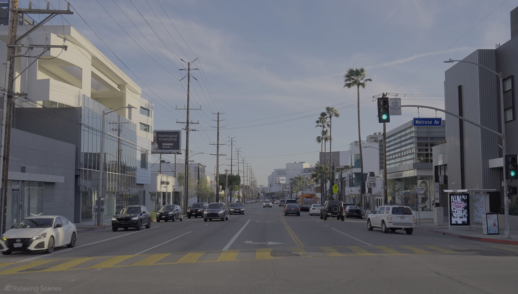} &
        \includegraphics[width=0.30\linewidth]{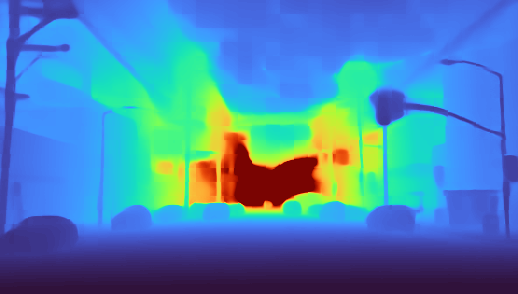} &
        \includegraphics[width=0.30\linewidth]{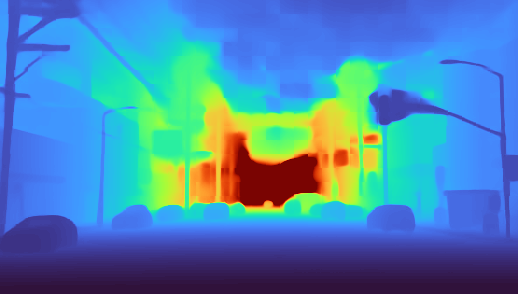} \\[4pt]

        \rotatebox{90}{~~~~~~~~~~~\textcolor{blue}{\textbf{Frame 2}}} &
        \includegraphics[width=0.30\linewidth]{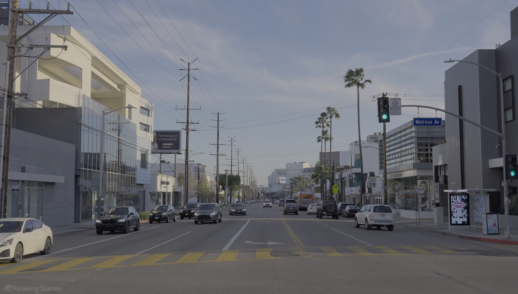} &
        \includegraphics[width=0.30\linewidth]{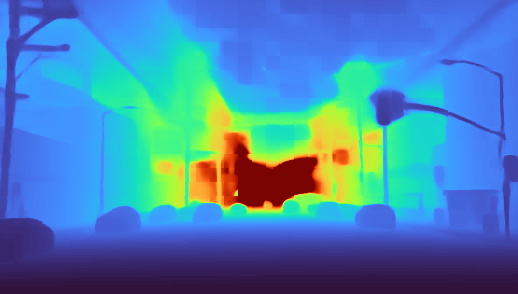} &
        \includegraphics[width=0.30\linewidth]{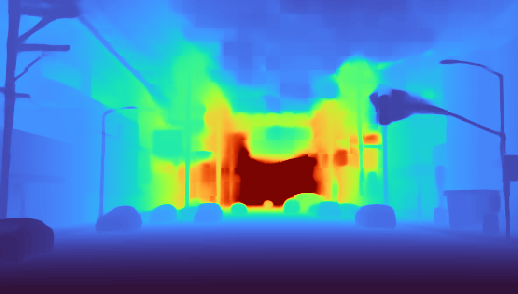} \\[4pt]

        \rotatebox{90}{~~~~~~~~~~~\textcolor{blue}{\textbf{Frame 3}}} &
        \includegraphics[width=0.30\linewidth]{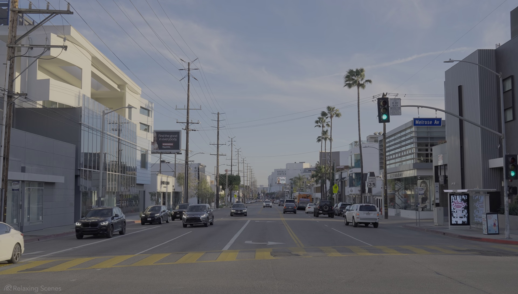} &
        \includegraphics[width=0.30\linewidth]{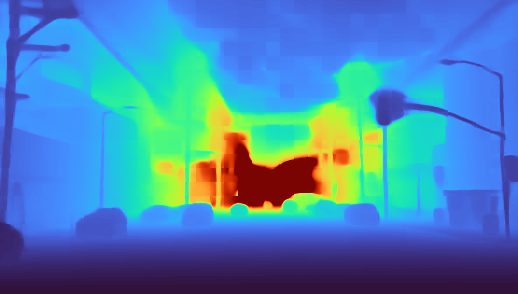} &
        \includegraphics[width=0.30\linewidth]{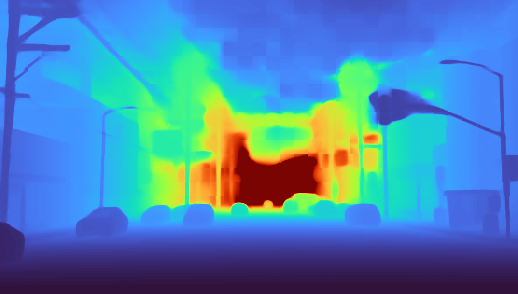} \\[4pt]

        \rotatebox{90}{~~~~~~~~~~~\textcolor{red}{\textbf{Frame 4}}} &
        \includegraphics[width=0.30\linewidth]{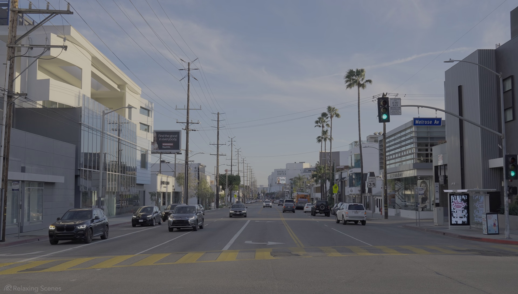} &
        \begin{tikzpicture}
          \node[inner sep=0pt] (img)
        {\includegraphics[width=0.30\linewidth]{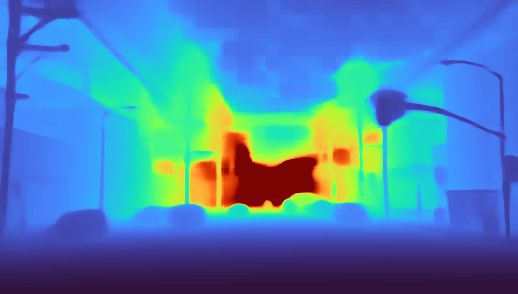}};
          \draw[red, dashed, ultra thick] (img.south west) rectangle (img.north east);
        \end{tikzpicture} &
        \includegraphics[width=0.30\linewidth]{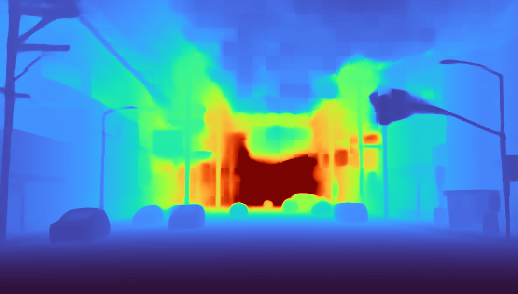} \\[4pt]

        \rotatebox{90}{~~~~~~~~~~~\textcolor{red}{\textbf{Frame 5}}} &
        \includegraphics[width=0.30\linewidth]{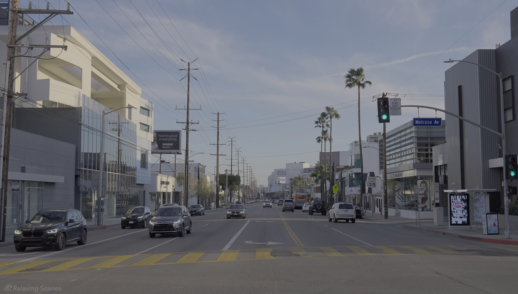} &
        \begin{tikzpicture}
          \node[inner sep=0pt] (img)
        {\includegraphics[width=0.30\linewidth]{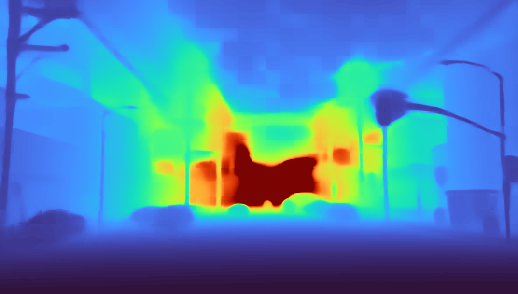}};
          \draw[red, dashed, ultra thick] (img.south west) rectangle (img.north east);
        \end{tikzpicture} &
        \includegraphics[width=0.30\linewidth]{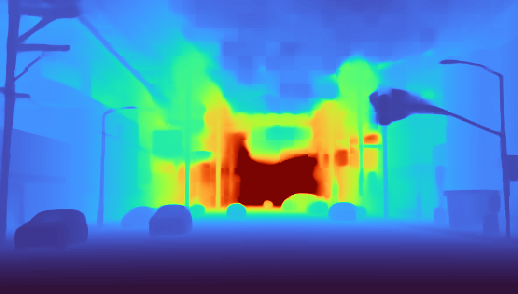} \\[4pt]

        \rotatebox{90}{~~~~~~~~~~~\textcolor{red}{\textbf{Frame 6}}} &
        \includegraphics[width=0.30\linewidth]{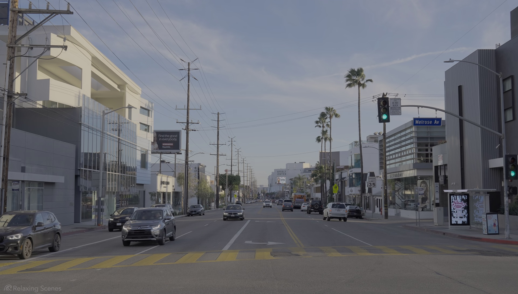} &
        \begin{tikzpicture}
          \node[inner sep=0pt] (img)
        {\includegraphics[width=0.30\linewidth]{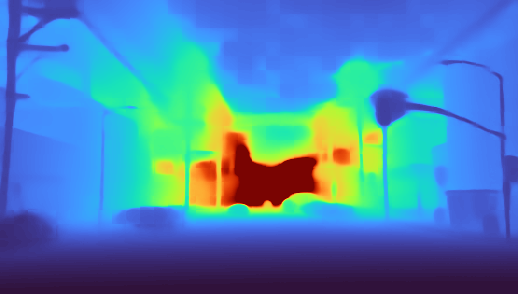}};
          \draw[red, dashed, thick] (img.south west) rectangle (img.north east);
        \end{tikzpicture} &
        \includegraphics[width=0.30\linewidth]{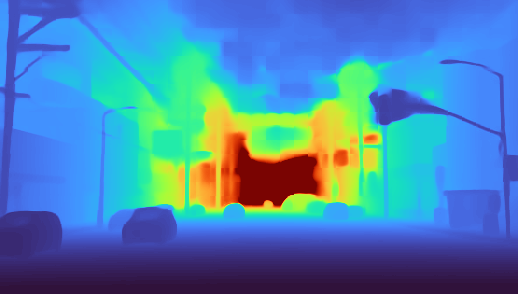} \\
    \end{tabular}

    \caption{Qualitative comparison of depth prediction for six frames. LFG is able to decouple static and dynamic objects as it continues along the road, and future work will improve the sharpness of the last frames' predictions. \textbf{Dashed red} outlines denote predicted frames with \emph{no ground-truth image input}, produced solely from the model’s future tokens}
\end{figure*}

\begin{figure*}[t]
    \centering
    \begin{tabular}{c c c c}
        & \textbf{RGB} & \textbf{LFG Depth (3 Frames)} & \textbf{ $\pi^3$ Depth (All Frames)} \\

        \rotatebox{90}{~~~~~~~~~~~\textcolor{blue}{\textbf{Frame 1}}} &
        \includegraphics[width=0.30\linewidth]{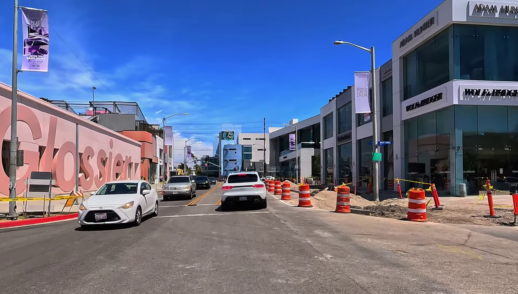} &
        \includegraphics[width=0.30\linewidth]{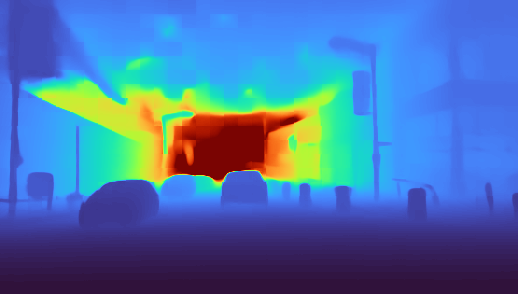} &
        \includegraphics[width=0.30\linewidth]{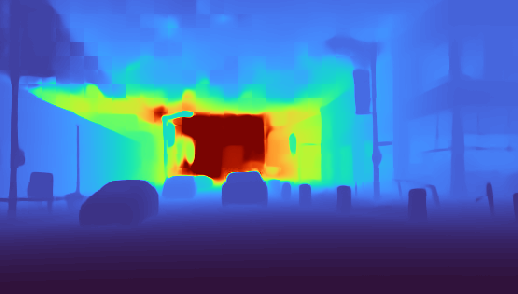} \\[4pt]

        \rotatebox{90}{~~~~~~~~~~~\textcolor{blue}{\textbf{Frame 2}}} &
        \includegraphics[width=0.30\linewidth]{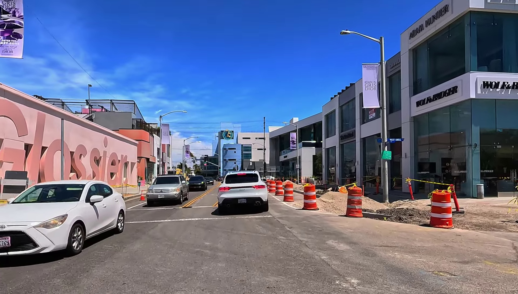} &
        \includegraphics[width=0.30\linewidth]{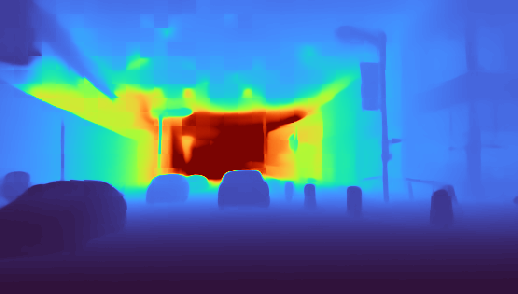} &
        \includegraphics[width=0.30\linewidth]{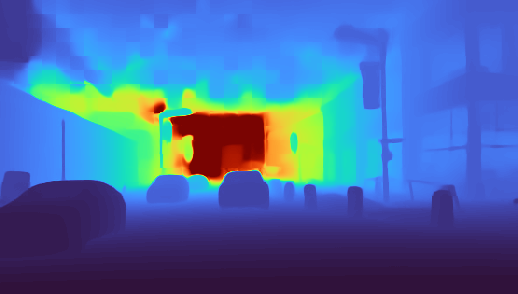} \\[4pt]

        \rotatebox{90}{~~~~~~~~~~~\textcolor{blue}{\textbf{Frame 3}}} &
        \includegraphics[width=0.30\linewidth]{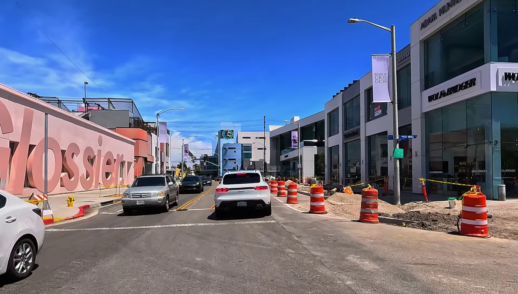} &
        \includegraphics[width=0.30\linewidth]{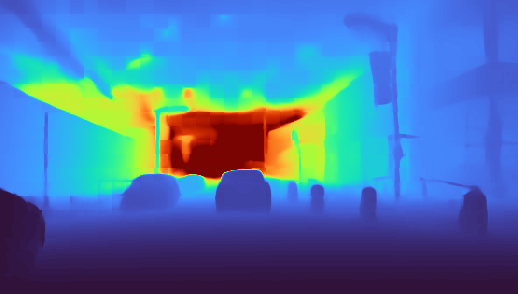} &
        \includegraphics[width=0.30\linewidth]{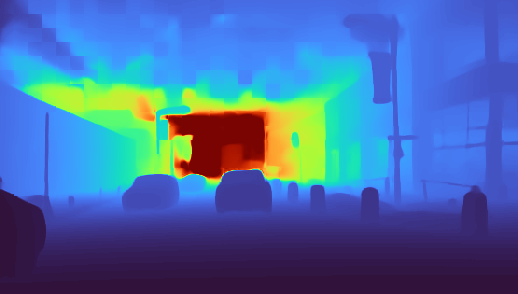} \\[4pt]

        \rotatebox{90}{~~~~~~~~~~~\textcolor{red}{\textbf{Frame 4}}} &
        \includegraphics[width=0.30\linewidth]{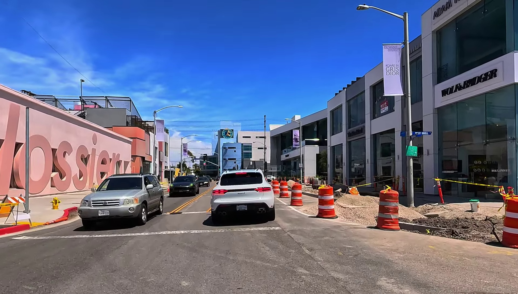} &
        \begin{tikzpicture}
          \node[inner sep=0pt] (img)
        {\includegraphics[width=0.30\linewidth]{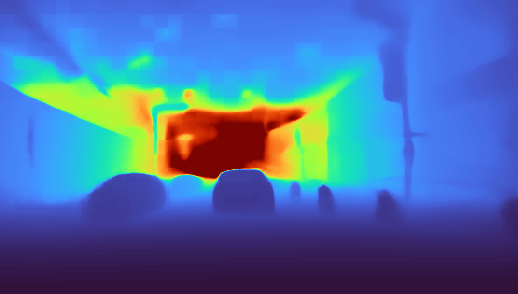}};
          \draw[red, dashed, ultra thick] (img.south west) rectangle (img.north east);
        \end{tikzpicture} &
        \includegraphics[width=0.30\linewidth]{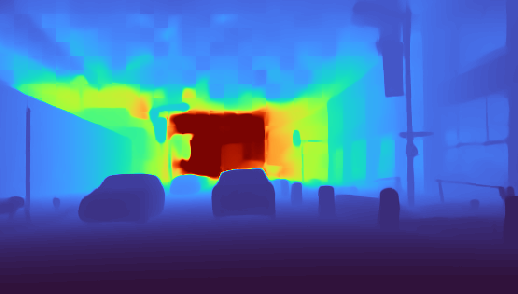} \\[4pt]

        \rotatebox{90}{~~~~~~~~~~~\textcolor{red}{\textbf{Frame 5}}} &
        \includegraphics[width=0.30\linewidth]{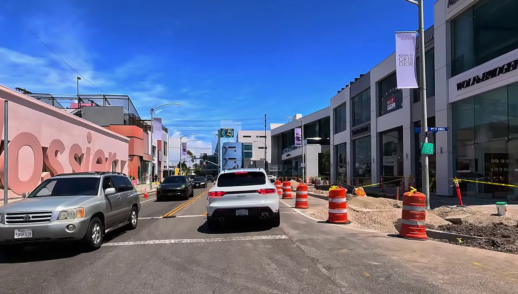} &
        \begin{tikzpicture}
          \node[inner sep=0pt] (img)
        {\includegraphics[width=0.30\linewidth]{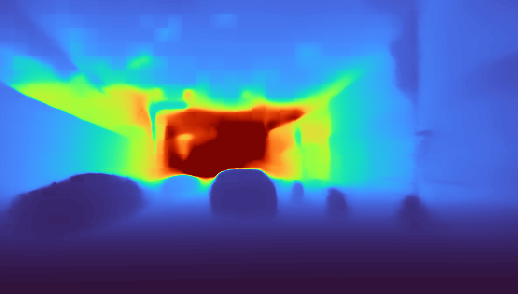}};
          \draw[red, dashed, ultra thick] (img.south west) rectangle (img.north east);
        \end{tikzpicture} &
        \includegraphics[width=0.30\linewidth]{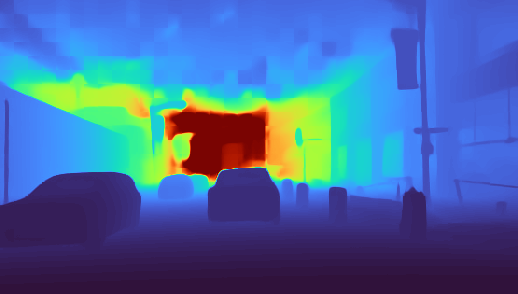} \\[4pt]

        \rotatebox{90}{~~~~~~~~~~~\textcolor{red}{\textbf{Frame 6}}} &
        \includegraphics[width=0.30\linewidth]{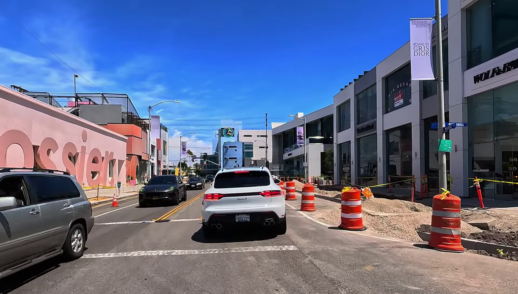} &
         \begin{tikzpicture}
          \node[inner sep=0pt] (img)
        {\includegraphics[width=0.30\linewidth]{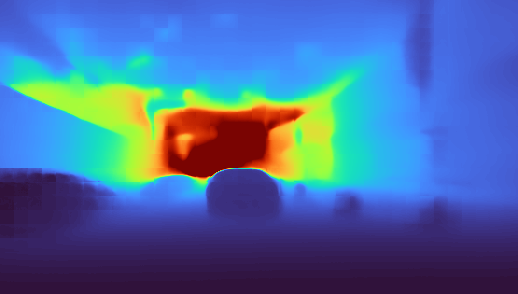}};
          \draw[red, dashed, thick] (img.south west) rectangle (img.north east);
        \end{tikzpicture} &
        \includegraphics[width=0.30\linewidth]{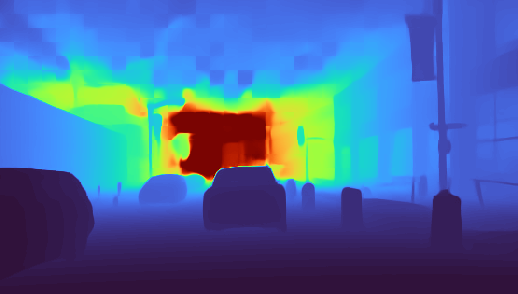} \\
    \end{tabular}

    \caption{More qualitative comparison of depth prediction for six frames. \textbf{Dashed red} outlines denote predicted frames with \emph{no ground-truth image input}, produced solely from the model’s future tokens}
\end{figure*}

\end{document}